\documentclass[11pt]{article}
\PassOptionsToPackage{hypertexnames=false}{hyperref}
\usepackage[preprint]{acl}
\usepackage{fix-cm}
\usepackage{fontspec}
\usepackage{amsmath,amssymb}
\usepackage{graphicx,booktabs,array,longtable,adjustbox}
\usepackage{enumitem,caption,placeins,fancyvrb,xurl,float,needspace,pdflscape}
\usepackage{microtype}

\makeatletter
\renewcommand{\small}{\@setfontsize\small{10pt}{12pt}}
\renewcommand{\footnotesize}{\@setfontsize\footnotesize{10pt}{12pt}}
\renewcommand{\Large}{\@setfontsize\Large{14pt}{16pt}}
\makeatother

\hypersetup{colorlinks=true,linkcolor=black,urlcolor=blue,citecolor=black,
pdftitle={MGE-PO: Entropy-Aware Credit Assignment for Mathematical Reasoning},
pdfauthor={Jihao Huang, Guangjia Chai, Zhirui Chen, Lei Wang, Yao Liu, Liu Qi, Pengjie Ding, Xinyang Liu, Chengxun Jia, Ling Shao, Hongyu Chen, Junchen Wan}}
\newcolumntype{L}[1]{>{\raggedright\arraybackslash}p{#1}}
\newcolumntype{C}[1]{>{\centering\arraybackslash}p{#1}}
\newlength{\tablecontentwidth}
\setlist{nosep,leftmargin=*}
\providecommand{\tightlist}{\setlength{\itemsep}{0pt}\setlength{\parskip}{0pt}}

\title{MGE-PO: Entropy-Aware Credit Assignment for Mathematical Reasoning}
\author{
\textbf{Jihao Huang \quad Guangjia Chai \quad Zhirui Chen \quad Lei Wang} \\
\textbf{Yao Liu \quad Liu Qi \quad Pengjie Ding \quad Xinyang Liu} \\
\textbf{Chengxun Jia \quad Ling Shao \quad Hongyu Chen \quad Junchen Wan} \\[0.4em]
{\normalfont\normalsize\href{mailto:jihao@cylingo.com}{\textcolor{black}{jihao@cylingo.com}}}
}
\date{}
\begin{document}

\maketitle
\begin{abstract}
GRPO assigns the same group-relative advantage to equally rewarded responses within a question, leaving their generation uncertainty unused. We introduce MGE-PO, centered on four-quadrant (4Q) advantage shaping that combines verified correctness with rollout generation entropy. Correctness determines the sign of an additional correction, while entropy controls its magnitude: lower-entropy successes receive larger bonuses and higher-entropy failures receive larger penalties. An online entropy-based question curriculum supports this credit rule using cached rollout measurements. At the final checkpoint after eight training epochs on Qwen3-8B-Base, adding 4Q across three policy optimization settings yields absolute gains of 3.23--3.69\% in six-benchmark macro avg@32 and 4.24--11.06\% in macro pass@32. On Qwen2.5-7B-Base, the complete configuration raises four-benchmark macro avg@32 from 41.5\% to 45.4\%.

\end{abstract}
\begin{figure*}[t]
\centering
\includegraphics[width=\linewidth]{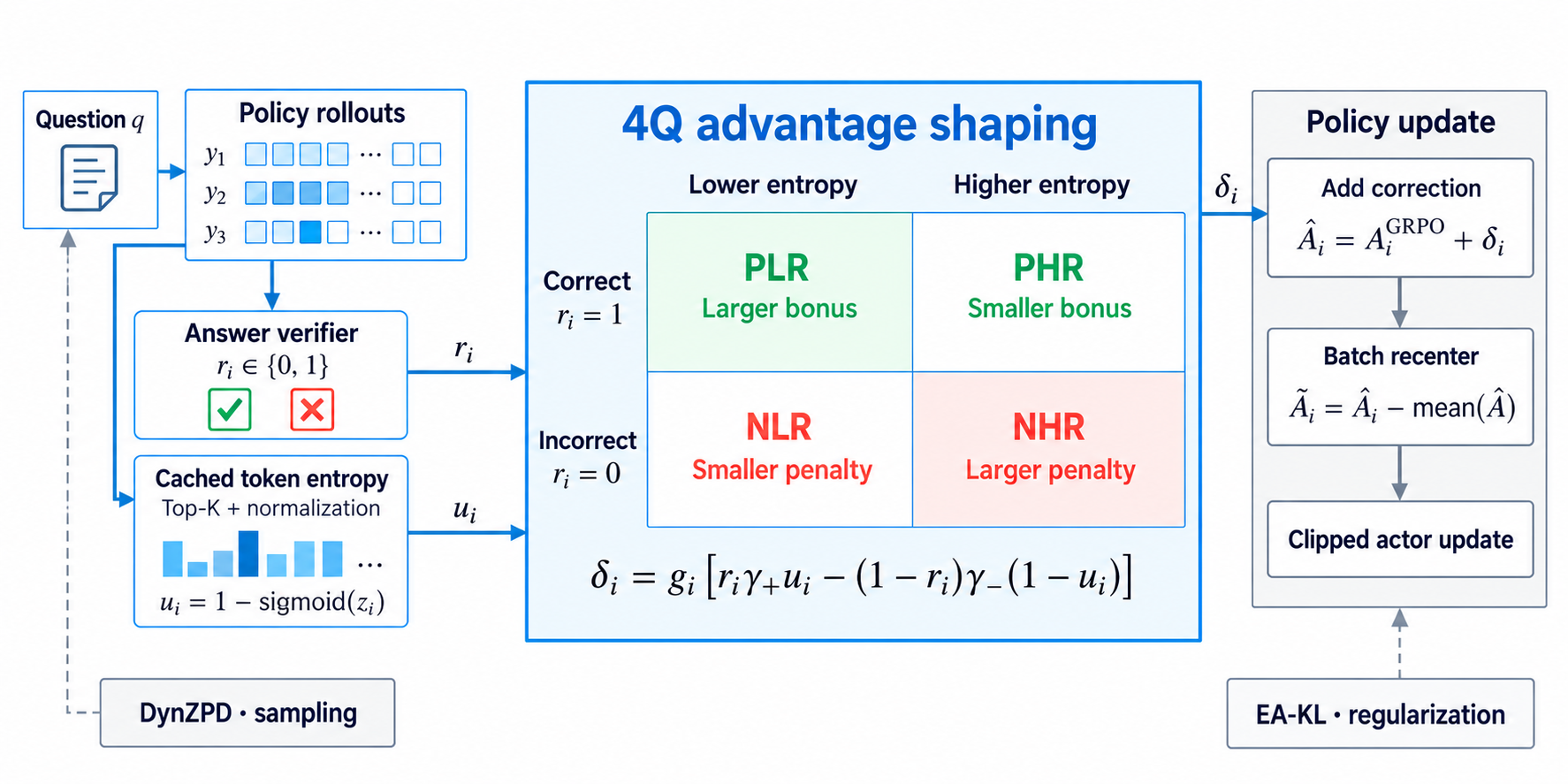}
\caption{GRPO-based MGE-PO centered on 4Q advantage shaping. Verified outcomes and cached token entropy determine continuous corrections: lower-entropy successes receive larger bonuses, higher-entropy failures larger penalties. Top-\(K\) pooling and running normalization produce \(u_i\). Corrections precede batch recentering and the clipped update; \(g_i=0\) disables them for truncated rollouts. High-entropy successes and low-entropy failures remain active. DynZPD and EA-KL support the complete configuration.}\label{fig:1}
\end{figure*}

\section{Introduction}

Reinforcement learning from verifiable rewards (RLVR) improves language-model reasoning using outcome feedback. For mathematical problems, an answer grader supplies a correctness reward without a learned reward model or step-level annotation. Group Relative Policy Optimization (GRPO) estimates advantages from groups of sampled answers and avoids a separate value network (Shao et al., 2024). DeepSeek-R1 and DAPO demonstrate the potential of this training paradigm (Guo et al., 2025; Yu et al., 2025).

Binary outcome feedback leaves a credit-assignment decision unresolved. In GRPO, responses to the same question with the same reward receive the same base advantage, even when the policy is concentrated along one response and dispersed along another. Entropy-conditioned updates can supplement outcome feedback, but differ in the uncertainty measure, unit of credit, and mapping from uncertainty to update strength (Cheng et al., 2026; Chen et al., 2025; He et al., 2026, v1). This motivates a specific question: how should verified outcomes and rollout generation uncertainty jointly determine additional credit?

We introduce MGE-PO, centered on entropy-aware advantage shaping (EA-Adv), which we call the four-quadrant (4Q) rule. The rule combines verified correctness with normalized rollout entropy in a continuous correction. Correctness selects its sign; entropy sets its magnitude. Lower-entropy successes receive larger additional bonuses, while higher-entropy failures receive larger additional penalties. Entropy measures predictive concentration along sampled prefixes, and the verifier supplies correctness. Conditioning on the outcome therefore allows concentration to guide credit even when a low-entropy response is incorrect. Cached token entropies, pooled over the most uncertain positions, provide the rollout statistic.

DynZPD supports 4Q by allocating questions through an online intermediate-entropy window with a positive sampling floor. It changes rollout opportunities, while 4Q changes rollout credit. Entropy-weighted reference KL is an actor optimization setting specified in Section 4.4 and Appendix B.4.

Our analysis characterizes the correction's sign and the within-outcome coefficient ordering preserved by batch recentering under binary-reward GRPO. A separate mixed-reward calculation motivates attention to partially mastered questions, with the entropy window assessed empirically.

On Qwen3-8B-Base, adding 4Q to GRPO yields absolute gains of 3.69\% in six-benchmark macro avg@32 and 11.06\% in macro pass@32 at the final checkpoint after eight epochs. Adding 4Q to DAPO and Clip-Cov also improves both macro metrics. The complete MGE-PO configuration raises four-benchmark macro avg@32 from 41.5\% to 45.4\% on Qwen2.5-7B-Base, with gains across the evaluated Qwen2.5 base initializations. The paired Qwen3 comparisons evaluate the credit rule, and the Qwen2.5 comparisons establish the performance of the complete configuration.

Our contributions are:

\begin{enumerate}
\def\labelenumi{\arabic{enumi}.}
\tightlist
\item
  An outcome-conditioned rollout advantage correction that uses generation entropy to distinguish responses with the same verified outcome, with relative-credit ordering characterized under binary-reward GRPO and batch recentering. An online entropy-based question curriculum supports this credit rule using cached training rollouts and a positive sampling floor.
\item
  Improvements in both average response accuracy and sampled question coverage from adding 4Q to three optimization settings on Qwen3-8B-Base, together with complete-configuration gains across the evaluated Qwen2.5 base initializations.
\end{enumerate}

\section{Related Work}

\textbf{Uncertainty and supervision in mathematical generation.} Malinin and Gales (2021) study token- and sequence-level uncertainty; Kuhn et al.~(2023) group meaning-equivalent answers to estimate semantic uncertainty. Token predictive entropy measures concentration at generated prefixes, whereas semantic entropy measures uncertainty across answers. MGE-PO pools token entropy for rollout credit and question sampling, supervised by verified final answers. Unlike human step-level reward supervision (Lightman et al., 2024) or Math-Shepherd's automatic process supervision (Wang et al., 2024), 4Q needs neither step-correctness labels nor a process reward model.

\textbf{Policy optimization and entropy-conditioned credit.} GRPO derives advantages from within-question rewards (Shao et al., 2024). DAPO adds asymmetric clipping, dynamic sampling of nondegenerate groups, token-level loss aggregation, and overlong reward shaping (Yu et al., 2025). EntroAdv adds a clipped, detached, nonnegative token-entropy bonus (Cheng et al., 2026); SEED-GRPO attenuates question-level updates using answer semantic entropy (Chen et al., 2025). EAPO v1 crosses advantage polarity with token entropy and amplifies high-entropy updates using group-normalized token entropy (He et al., 2026, v1). MGE-PO instead uses verified correctness to select a rollout correction's sign: lower entropy increases the success bonus, and higher entropy increases the failure penalty. This coefficient enters the rollout advantage. The distinctions concern the credit unit, outcome signal, and within-outcome entropy mapping.

\textbf{Adaptive curricula.} Competence-based curricula expand the difficulty range during learning (Platanios et al., 2019); uncertainty-aware translation curricula combine difficulty with model uncertainty (Zhou et al., 2020). SENT uses initial-policy semantic entropy for easy-to-hard reasoning stages (Cao et al., 2025). EDCO periodically estimates entropy from short prefixes and selects high-entropy subsets, also testing a moderate-entropy window (Pang et al., 2026). DynZPD instead updates question statistics from training rollouts and samples through a bell-shaped distribution with a positive floor, supplying opportunities for rollout credit assignment.

\textbf{Reference regularization in RLVR.} Selective KL penalties vary reference costs by token properties (Vassoyan et al., 2025; Cui et al., 2025; Cao et al., 2025), while DAPO omits frozen-reference KL (Yu et al., 2025). MGE-PO uses bounded entropy-dependent token weights. Appendices B.4 and D.7 specify this optimization setting and its diagnostics; Appendix E compares intervention rules.

\section{Preliminaries}

\subsection{Group-relative outcome supervision}

For question \(q\), policy \(\pi_k\) samples \(y_i=(a_{i,1},\ldots,a_{i,T_i})\). A deterministic final-answer grader gives 4Q binary correctness \(r_i\in\{0,1\}\); host reward shaping may separately determine the base advantage.

Binary-reward GRPO normalizes \(G\) completions of \(q\):

\[
\adjustbox{max width=\dimexpr\linewidth-24pt\relax}{$\displaystyle \begin{gathered}
A_i^{\mathrm{GRPO}}
=\frac{r_i-\mu_q}{\sigma_q+\epsilon},\\
\mu_q=\frac{1}{G}\sum_{j=1}^{G}r_j,
\end{gathered}$}
\]

Here \(\sigma_q\) is the empirical reward standard deviation and \(\epsilon>0\). Equal rewards give equal coefficients; homogeneous groups give zero.

At prefix \(c_{i,t}=(q,a_{i,<t})\), the likelihood ratio and clipped token surrogate are

\[
\adjustbox{max width=\dimexpr\linewidth-24pt\relax}{$\displaystyle \rho_{i,t}(\theta)
=\frac{\pi_\theta(a_{i,t}\mid c_{i,t})}
{\pi_k(a_{i,t}\mid c_{i,t})},$}
\]

\[
\adjustbox{max width=\dimexpr\linewidth-24pt\relax}{$\displaystyle \begin{aligned}
\ell_{i,t}^{\mathrm{clip}}(\theta,A_i)
&=\min\Bigl[\rho_{i,t}(\theta)A_i,\\
&\quad\operatorname{clip}\bigl(\rho_{i,t}(\theta),
1-\epsilon_\ell,1+\epsilon_h\bigr)A_i\Bigr].
\end{aligned}$}
\]

Matched component comparisons use the same masked actor-batch aggregation \(\mathcal L_{\mathrm{clip}}\). The rollout policy \(\pi_k\) is refreshed between rounds; fixed \(\pi_{\mathrm{ref}}\) defines the reference-deviation cost.

\subsection{Generation entropy}

Generation entropy is the next-token Shannon entropy:

\[
\adjustbox{max width=\dimexpr\linewidth-24pt\relax}{$\displaystyle H_{i,t}
=-\sum_{v\in\mathcal V}
\pi_k(v\mid c_{i,t})\log\pi_k(v\mid c_{i,t}).$}
\]

Here \(\pi_k\) is the pre-decoding policy distribution; entropy is in nats and excludes padding. Cached entropy statistics and coefficients are detached during actor updates. Entropy measures concentration, separately from correctness; Appendix C.1 lists the information needed to reproduce the cached entropy measurements.

\section{MGE-PO}

MGE-PO uses 4Q to assign rollout credit, DynZPD to sample questions, and reference regularization in the actor objective (Figure 1).

\subsection{Entropy summaries for rollout and question decisions}

Rollout credit pools the highest-entropy positions:

\[
\adjustbox{max width=\dimexpr\linewidth-24pt\relax}{$\displaystyle \begin{gathered}
h_i^{\mathrm{top}}
=\frac{1}{K_i}
\sum_{t\in\operatorname{TopK}(H_{i,\cdot},K_i)}H_{i,t},\\
K_i=\min\bigl(T_i,\max(1,\min(\lfloor r_KT_i\rfloor,K_{\max}))\bigr).
\end{gathered}$}
\]

With \(r_K=0.1\) and \(K_{\max}=4096\), running mean \(\mu_H\) and scale \(\tau_H\) normalize this uncertainty summary:

\[
\adjustbox{max width=\dimexpr\linewidth-24pt\relax}{$\displaystyle \begin{gathered}
z_i=\frac{h_i^{\mathrm{top}}-\mu_H}{\tau_H+\epsilon_u},\\
u_i=1-\operatorname{sigmoid}(z_i).
\end{gathered}$}
\]

Higher \(u_i\) means lower relative pooled entropy. Appendix B.1 specifies the EMA with \(\lambda_H=0.99\) and \(\epsilon_u=10^{-6}\).

Question sampling uses all active generated tokens:

\[
\adjustbox{max width=\dimexpr\linewidth-24pt\relax}{$\displaystyle h_i^{\mathrm{mean}}=\frac{1}{T_i}\sum_{t=1}^{T_i}H_{i,t}.$}
\]

Both summaries reuse cached token entropies: top-entropy pooling targets uncertain prefixes, while the full-token mean describes overall generation uncertainty.

\subsection{4Q: outcome-conditioned advantage shaping}

The 4Q rule, also denoted EA-Adv in the component experiments, adds an outcome-conditioned correction to a base advantage coefficient. Its mathematical interface is

\[
\adjustbox{max width=\dimexpr\linewidth-24pt\relax}{$\displaystyle \begin{gathered}
\widehat A_i=A_i^{\mathrm{base}}+\delta_i,\\
\delta_i=g_i\left[
r_i\gamma_+u_i-(1-r_i)\gamma_-(1-u_i)
\right],\\
g_i=1-\mathbf{1}[y_i\text{ is truncated}].
\end{gathered}$}
\]

For the GRPO instantiation, \(A_i^{\mathrm{base}}=A_i^{\mathrm{GRPO}}\), giving \(\widehat A_i=A_i^{\mathrm{GRPO}}+\delta_i\). Other host optimizers supply their own base coefficients, while 4Q uses the binary verification outcome independently of any shaped reward. Appendix C.4 records the integration scope for the DAPO and Clip-Cov combinations.

Correctness determines the sign of the direct correction, and normalized rollout entropy determines its magnitude. Lower-entropy successes receive larger additional bonuses; higher-entropy failures receive larger additional penalties. This preference emphasizes successful behavior with concentrated generation distributions and downweights failed behavior with dispersed distributions. The Qwen2.5 configuration uses \(\gamma_+=0.5\) and \(\gamma_-=0.3\); Appendix C.4 describes the separate Qwen3 training and evaluation protocol.

The four-quadrant terminology crosses verified correctness with normalized rollout entropy: PLR denotes low-entropy correct rollouts, PHR high-entropy correct rollouts, NLR low-entropy incorrect rollouts, and NHR high-entropy incorrect rollouts. These labels describe rollout regions; the sigmoid interpolates continuously within each outcome class. The default combines a concentration-weighted positive term and an uncertainty-weighted negative term, so high-entropy successes and low-entropy failures still receive continuous corrections. Appendix B.2 introduces separately weighted PHR and NLR basis terms for the coefficient comparisons.

Figure 2 visualizes this continuous correction within each outcome class.

\begin{figure}[t]
\centering
\includegraphics[width=\linewidth]{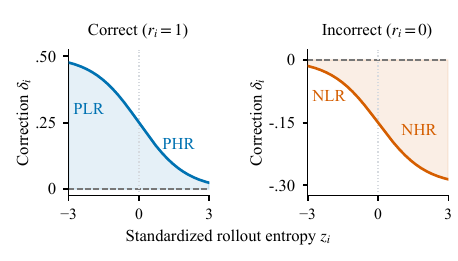}
\caption{Continuous 4Q corrections for \(g_i=1\), \(\gamma_+=0.5\), and \(\gamma_-=0.3\) (Qwen2.5). The horizontal axis is standardized rollout entropy \(z_i\) (Section 4.1). Lower entropy increases the success bonus; higher entropy increases the failure penalty. PLR/PHR and NLR/NHR mark within-outcome entropy regions; dashed lines mark zero. Corrections precede batch recentering, and PHR/NLR remain active.}\label{fig:2}
\end{figure}

A fixed outcome-dependent correction changes the separation between successes and failures. Entropy additionally distinguishes equally rewarded responses within a question, giving the relative-credit ordering analyzed in Section 5.2.

The specified GRPO-based MGE-PO actor uses batch-recentered advantages:

\[
\adjustbox{max width=\dimexpr\linewidth-24pt\relax}{$\displaystyle \widetilde A_i
=\widehat A_i-\frac{1}{M}\sum_{j=1}^{M}\widehat A_j,$}
\]

where \(M\) counts the rollouts included in recentering. This places the batch at zero mean and expresses each coefficient relative to the other rollouts. The sign interpretation above applies to \(\delta_i\); \(\widetilde A_i\) includes the base advantage and shared centering shift. The truncation gate removes the direct correction before centering. Appendix B.3 describes truncation handling and the resulting behavior in homogeneous reward groups.

\subsection{DynZPD: online question sampling}

For each question's rollout group \(\mathcal B_q\), DynZPD computes

\[
\adjustbox{max width=\dimexpr\linewidth-24pt\relax}{$\displaystyle e_q^{(k)}
=\frac{1}{|\mathcal B_q|}\sum_{i\in\mathcal B_q}h_i^{\mathrm{mean}}.$}
\]

Its stored statistic follows the EMA

\[
\adjustbox{max width=\dimexpr\linewidth-24pt\relax}{$\displaystyle \xi_q^{(k)}
=\lambda_q\xi_q^{(k-1)}+(1-\lambda_q)e_q^{(k)}.$}
\]

Unobserved questions retain their statistics. Round \(k\) samples using values from round \(k-1\):

\[
\adjustbox{max width=\dimexpr\linewidth-24pt\relax}{$\displaystyle \begin{gathered}
v_q^{(k)}
=\eta+(1-\eta)
\exp\left[-\frac{(\xi_q^{(k-1)}-\mu_{\mathrm{ZPD}})^2}
{2\sigma_{\mathrm{ZPD}}^2}\right],\\
p_k(q)=\frac{v_q^{(k)}}{\sum_{q'\in\mathcal D}v_{q'}^{(k)}}.
\end{gathered}$}
\]

The Gaussian favors the target entropy; \(\eta\) retains positive dataset-wide support. Settings are \(\mu_{\mathrm{ZPD}}=0.30\), \(\sigma_{\mathrm{ZPD}}=0.15\), \(\eta=0.10\), and \(\lambda_q=0.9\). Initializing \(\xi_q^{(0)}=\mu_{\mathrm{ZPD}}\) gives uniform sampling.

This model-relative curriculum changes training emphasis. Section 5 relates partial mastery to learning opportunities; Section 6 reports allocation diagnostics. Appendix B.5 details the sampling floor and statistic updates.

\subsection{Actor optimization}

The GRPO-based actor optimizes recentered advantages on curriculum-sampled questions:

\[
\adjustbox{max width=\dimexpr\linewidth-24pt\relax}{$\displaystyle \begin{aligned}
J_k(\theta)&=\mathbb E_{\mathcal B_k\sim(p_k,\pi_k)}\Bigl[
\mathcal L_{\mathrm{clip}}(\theta;\widetilde{\mathbf A})\\
&\hspace{15mm}-\beta\widehat{\mathcal K}_w(\theta,\pi_{\mathrm{ref}})\Bigr].
\end{aligned}$}
\]

EA-KL uses detached token weights: reference deviation costs more below the batch mean entropy and less above it. The complete Qwen2.5 configuration uses \(\beta=0.001\), \(\alpha=0.20\); its GRPO baseline uses \(\beta=0.04\) (Appendix D.10). Setting \(\alpha=0\) retains uniform KL, while \(\beta=0\) removes it. Appendix B.4 specifies weights, masks, and the sampled penalty.

Entropy coefficients stay fixed within each actor update; the rollout policy refreshes between rounds. Appendices B.6 and C give the procedure and configurations.

\section{Properties of Outcome-Conditioned Entropy Shaping}

Entropy statistics are frozen within each update. Section 5.1 concerns direct corrections; Section 5.2 assumes binary-reward GRPO normalization and shared batch recentering. Appendix A gives the sampling proof and idealized optimization interpretation.

\subsection{Outcome-conditioned direction and magnitude}

For \(\gamma_+,\gamma_-\ge0\), \(u_i\in[0,1]\), and \(g_i\in\{0,1\}\), the direct correction satisfies

\begin{equation}
\adjustbox{max width=\dimexpr\linewidth-36pt\relax}{$\displaystyle \begin{gathered}
-g_i\gamma_-\le\delta_i\le g_i\gamma_+,\\
r_i=1\Rightarrow\delta_i\ge0,\\
r_i=0\Rightarrow\delta_i\le0.
\end{gathered}$}
\tag{S1}\end{equation}

At fixed running statistics, \(u_i=1-\operatorname{sigmoid}(z_i)\) decreases with entropy:

\[
\adjustbox{max width=\dimexpr\linewidth-24pt\relax}{$\displaystyle \frac{\partial\delta_i}{\partial z_i}
=-g_i\left[r_i\gamma_++(1-r_i)\gamma_-\right]u_i(1-u_i)\le0.$}
\]

For untruncated responses and positive default coefficients, entropy reduces the success bonus or increases the failure penalty. This bounded, continuous rule assigns sign by correctness and strength by concentration. The derivative concerns cached inputs; entropy remains detached during actor updates.

\subsection{Relative credit survives batch recentering}

\textbf{Property 1 (Relative-credit ordering).} Consider two untruncated responses \(i,j\) from the same binary-reward GRPO normalization group and actor batch, with the same outcome and shared running entropy statistics. Their base GRPO advantages are equal. With \(a_1=\gamma_+\) and \(a_0=\gamma_-\), batch recentering gives

\begin{equation}
\adjustbox{max width=\dimexpr\linewidth-36pt\relax}{$\displaystyle \widetilde A_i-\widetilde A_j
=\delta_i-\delta_j
=a_{r_i}(u_i-u_j).$}
\tag{S2}\end{equation}

Equal base advantages and the shared centering constant cancel. For positive coefficients, lower entropy gives a larger success bonus or a smaller failure penalty within the same outcome class.

Replacing \(u_i\) by \(u_0\in[0,1]\) gives the untruncated outcome-only correction \(\delta_i^{\mathrm{const}}=r_i\gamma_+u_0-(1-r_i)\gamma_-(1-u_0)\) and decomposition

\begin{equation}
\adjustbox{max width=\dimexpr\linewidth-36pt\relax}{$\displaystyle \delta_i
=\delta_i^{\mathrm{const}}
+a_{r_i}(u_i-u_0).$}
\tag{S2a}\end{equation}

For the same-outcome pair in Property 1, the constant correction gives identical additional credit and zero coefficient difference after recentering. Equation (S2a) identifies the additional contribution of entropy: response-dependent coefficient variation with the ordering in (S2). A correction determined only by the outcome and shared group statistics changes the separation between outcome classes while leaving this same-outcome pair undifferentiated.

For a correct response \(c\) and an incorrect response \(f\) in the same binary-reward GRPO normalization group and actor batch, the same cancellation yields

\begin{equation}
\adjustbox{max width=\dimexpr\linewidth-36pt\relax}{$\displaystyle \begin{aligned}
\widetilde A_c-\widetilde A_f
&=\frac{1}{\sigma_q+\epsilon}
+g_c\gamma_+u_c+g_f\gamma_-(1-u_f)\\
&\ge A_c^{\mathrm{GRPO}}-A_f^{\mathrm{GRPO}}.
\end{aligned}$}
\tag{S3}\end{equation}

Under the specified binary-reward GRPO normalization, 4Q preserves within-outcome entropy ordering and weakly increases the correct-versus-incorrect coefficient gap. These are properties of relative credit; an individual centered advantage can have a different sign from its direct correction (Appendix B.3). The resulting 4Q training rule is evaluated in Section 6.2, with coefficient and component comparisons in Appendices D.6 and D.10.

\subsection{Sampling opportunities for outcome-relative learning}

For success probability \(s_q\), \(G\ge2\) independent completions give mixed-reward probability

\begin{equation}
\adjustbox{max width=\dimexpr\linewidth-36pt\relax}{$\displaystyle m_G(s_q)=1-s_q^G-(1-s_q)^G.$}
\tag{S4}\end{equation}

\textbf{Proposition 1 (Mixed-reward opportunities).} \(m_G\) is uniquely maximized at \(s_q=1/2\) and vanishes at \(s_q\in\{0,1\}\) (proof: Appendix A.1).

Mixed groups give nonzero base GRPO coefficients, motivating partially mastered questions. DynZPD's intermediate-entropy window is an empirical proxy for this success-probability regime, tested through component comparisons. 4Q can also differentiate responses within reward-homogeneous groups (Appendix B.3).

Shared parameters and a finite update budget make question allocation consequential. DynZPD changes gradient frequencies without inverse-probability correction (Appendix A.1); Section 6.4 reports sampling concentration.

\begin{table*}[t]
\centering
\caption{Qwen3-8B-Base at final controller step 624 (eight epochs). Cells report \textbf{avg@32 / pass@32 (\%)} using 32 responses per question, temperature 0.6, and matched test sets and decoding. Avg@32 equals mean@32. Macro scores weight six benchmarks equally. Bold marks each metric’s maximum across seven configurations, including ties. Displayed scores use two decimals; macro averages retain reported precision before rounding.}\label{tab:1}
\fontsize{10}{12}\selectfont
\setlength{\tabcolsep}{2.5pt}
\renewcommand{\arraystretch}{1.15}
\setlength{\tablecontentwidth}{\dimexpr\linewidth-35.0pt\relax}
\begin{tabular}{@{}L{0.17000\tablecontentwidth}C{0.11857\tablecontentwidth}C{0.11857\tablecontentwidth}C{0.11857\tablecontentwidth}C{0.11857\tablecontentwidth}C{0.11857\tablecontentwidth}C{0.11857\tablecontentwidth}C{0.11857\tablecontentwidth}@{}}
\toprule
\textbf{Method} & \textbf{AIME2025 (\%)} & \textbf{AIME2024 (\%)} & \textbf{AMC2023 (\%)} & \textbf{MATH500 (\%)} & \textbf{Minerva (\%)} & \textbf{Olympiad\newline  Bench (\%)} & \textbf{Macro (\%)} \\
\midrule
GRPO & 13.44 /\newline 33.33 & 5.52 /\newline 13.33 & 61.02 /\newline 85.00 & 76.45 /\newline 89.40 & 46.66 /\newline 60.29 & 44.78 /\newline 66.47 & 41.31 /\newline 57.97 \\
DAPO & 16.35 /\newline 40.00 & 10.73 /\newline 23.33 & 65.47 /\newline 92.50 & 78.79 /\newline 92.60 & 44.47 /\newline \textbf{74.63} & 45.87 /\newline 75.22 & 43.61 /\newline 66.38 \\
Clip-Cov & 15.52 /\newline 43.33 & 10.83 /\newline 33.33 & 63.59 /\newline 92.50 & 78.40 /\newline 93.20 & 45.23 /\newline 72.79 & 45.92 /\newline 72.85 & 43.25 /\newline 68.00 \\
MGE-PO & \textbf{23.96} /\newline 53.33 & 13.02 /\newline 40.00 & 64.53 /\newline \textbf{97.50} & 81.31 /\newline 93.00 & \textbf{47.48} /\newline 68.01 & 52.47 /\newline 74.33 & \textbf{47.13} /\newline 71.03 \\
Clip-Cov + 4Q & 20.42 /\newline \textbf{56.67} & 16.25 /\newline \textbf{43.33} & \textbf{67.81} /\newline 92.50 & 81.11 /\newline \textbf{94.00} & 44.16 /\newline 70.96 & 49.11 /\newline 75.96 & 46.48 /\newline \textbf{72.24} \\
4Q & 18.85 /\newline 46.67 & 12.60 /\newline 36.67 & 62.66 /\newline 92.50 & 79.66 /\newline 92.60 & 46.90 /\newline 69.49 & 49.33 /\newline 76.26 & 45.00 /\newline 69.03 \\
4Q + DAPO & 20.31 /\newline 50.00 & \textbf{16.67} /\newline 40.00 & 63.20 /\newline \textbf{97.50} & \textbf{82.54} /\newline 93.20 & 46.61 /\newline 70.96 & \textbf{52.62} /\newline \textbf{78.49} & 46.99 /\newline 71.69 \\
\bottomrule
\end{tabular}
\end{table*}

\section{Experiments}

Qwen3 comparisons test 4Q across three optimization settings; Qwen2.5 comparisons evaluate complete configurations. Diagnostics examine predictive concentration, training, and sampling.

\subsection{Experimental setup}

Qwen3-8B-Base (Yang et al., 2025) uses a random 10,000-question Skywork-OR1 mathematics subset (He et al., 2025), subsampled with seed 1; filtering leaves 9,998 questions. Batches contain 128 questions with eight rollouts each, capped at 20,480 tokens. All seven configurations are evaluated at final controller step 624 after eight epochs (Appendix C.4).

Learning rate, benchmark sets, and decoding settings are matched. Evaluation samples 32 responses per question at temperature 0.6 on AIME2025, AIME2024, AMC2023, MATH500, Minerva, and OlympiadBench. We denote mean@32 by avg@32; macro scores weight the six benchmarks equally.

The original series uses Qwen2.5-7B-Base, Qwen2.5-Math-1.5B-Base, Qwen2.5-Math-7B-Base, and an additional Math-7B-Instruct initialization (Yang et al., 2024a,b), trained on 17,398 DAPO-Math-17k questions (Yu et al., 2025). Base runs use eight epochs, batch size 128, 16 rollouts per question, and a 12,288-token maximum. Under adaptive sampling, epochs measure dataset-equivalent budgets, not guaranteed visits.

Qwen2.5 evaluation uses avg@32/pass@32 on MATH-500, AMC23, AIME24, and AIME25, with temperature 1.0, top-\(p\) 0.7, and approximately 12k tokens. Supplementary evaluation uses avg@64 on MATH-500, AMC23, Minerva-Math, and Olympiad, plus AIME24/AIME25 pass@256; Minerva/Olympiad responses are capped at 8,192 tokens. Protocols remain separate because inference budgets and checkpoint metadata differ. Table 9 gives sizes and metrics; MATH and Minerva follow Hendrycks et al.~(2021) and Lewkowycz et al.~(2022).

For \(N\) questions with \(n\) responses each and \(c_j\) correct responses on question \(j\),

\[
\adjustbox{max width=\dimexpr\linewidth-24pt\relax}{$\displaystyle \operatorname{avg}@n=\frac{100}{N}\sum_{j=1}^{N}\frac{c_j}{n}.$}
\]

Pass@\(K\) is the proportion of questions with at least one correct response among \(K\) samples (Chen et al., 2021), averaged over \(R_K\) evaluations of the same checkpoint:

\[
\adjustbox{max width=\dimexpr\linewidth-24pt\relax}{$\displaystyle \operatorname{pass}@K
=\frac{100}{N R_K}\sum_{\ell=1}^{R_K}\sum_{j=1}^{N}
\mathbf{1}\!\left[c_{j,\ell}^{(K)}>0\right].$}
\]

Here \(c_{j,\ell}^{(K)}\) counts correct responses among \(K\) samples in repeat \(\ell\); a single evaluation has \(R_K=1\). Qwen2.5 averages inference repeats of one training run per model-method cell; Qwen3 reports final-step point estimates. Neither measures training-seed variability (Appendix C).

Scores use percentages (\%); changes are absolute differences, \(\Delta S=S_{\mathrm{method}}-S_{\mathrm{baseline}}\), also denoted \%.

Qwen2.5 baselines are GRPO, DAPO, and KL-Cov, plus supplementary Clip-Cov on Minerva/Olympiad. Qwen3 compares GRPO, DAPO, and Clip-Cov on all six benchmarks. Appendix D.1 includes pre-RL Qwen2.5 scores.

\begin{figure}[t]
\centering
\includegraphics[width=\linewidth]{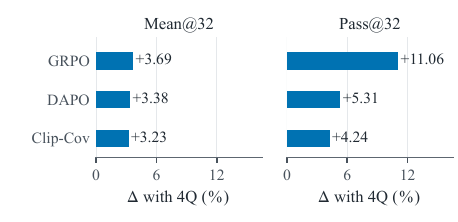}
\caption{Adding 4Q to GRPO, DAPO, and Clip-Cov on Qwen3-8B-Base at step 624. Bars show absolute macro mean@32 (avg@32) and pass@32 changes (\%), computed before rounding. Pairings are those in Section 6.2; values are point estimates without uncertainty intervals.}\label{fig:3}
\end{figure}

\subsection{4Q across optimization settings on Qwen3-8B-Base}

\begin{table*}[t]
\centering
\caption{Four-benchmark summary (\%): macro scores are unweighted means over MATH-500, AMC23, AIME24, and AIME25 (full scores: Appendix D.1). Each model-method cell uses one training run. MGE-PO-GRPO comparisons include a global KL-coefficient difference (Appendices C.3 and D.10).}\label{tab:2}
\fontsize{10}{12}\selectfont
\setlength{\tabcolsep}{4pt}
\renewcommand{\arraystretch}{1.15}
\setlength{\tablecontentwidth}{\dimexpr\linewidth-40pt\relax}
\begin{tabular}{@{}L{0.30000\tablecontentwidth}L{0.12000\tablecontentwidth}C{0.14500\tablecontentwidth}C{0.14500\tablecontentwidth}C{0.14500\tablecontentwidth}C{0.14500\tablecontentwidth}@{}}
\toprule
\textbf{Initialization} & \textbf{Method} & \textbf{Macro\newline  avg@32 (\%)} & \textbf{Macro\newline  pass@32 (\%)} & \textbf{AIME24\newline  pass@32 (\%)} & \textbf{AIME25\newline  pass@32 (\%)} \\
\midrule
Qwen2.5-Math-1.5B-Base & GRPO & 33.7 & 60.1 & 30.3 & 30.7 \\
 & DAPO & 35.3 & 59.2 & 32.7 & 26.7 \\
 & KL-Cov & 35.8 & 61.5 & 31.8 & 35.7 \\
 & MGE-PO & \textbf{36.0} & \textbf{63.3} & \textbf{33.3} & \textbf{38.3} \\
\addlinespace[3pt]
Qwen2.5-Math-7B-Base & GRPO & 41.4 & 63.2 & 36.7 & 35.4 \\
 & DAPO & 42.7 & 65.3 & 43.3 & 36.7 \\
 & KL-Cov & \textbf{43.6} & 66.2 & 44.4 & 38.3 \\
 & MGE-PO & 43.5 & \textbf{70.0} & \textbf{50.0} & \textbf{43.3} \\
\addlinespace[3pt]
Qwen2.5-7B-Base & GRPO & 41.5 & 58.6 & 30.3 & 23.3 \\
 & DAPO & 42.5 & 60.1 & 36.7 & 25.7 \\
 & KL-Cov & 44.4 & 62.2 & 39.5 & 28.3 \\
 & MGE-PO & \textbf{45.4} & \textbf{62.8} & \textbf{40.0} & \textbf{30.3} \\
\bottomrule
\end{tabular}
\end{table*}

Adding 4Q improves macro accuracy and coverage across all three optimization settings (Table 1; Figure 3). The paired configurations are GRPO versus 4Q, DAPO versus 4Q + DAPO, and Clip-Cov versus Clip-Cov + 4Q; Table 1 also reports complete MGE-PO.

GRPO's macro avg@32 rises from 41.31\% to 45.00\% and pass@32 from 57.97\% to 69.03\%. DAPO gains 3.38\%/5.31\%; Clip-Cov gains 3.23\%/4.24\%. Standalone 4Q also exceeds unaugmented DAPO and Clip-Cov on both macro metrics (paired differences: Appendix D.11).

Complete MGE-PO leads macro avg@32 at 47.13\% (+5.82\% over GRPO), with pass@32 of 71.03\% (+13.06\%). Clip-Cov + 4Q leads pass@32 at 72.24\%, followed by 4Q + DAPO at 71.69\%. Accuracy and coverage identify different leading configurations.

The macro gains include benchmark-specific tradeoffs. Adding 4Q to DAPO lowers AMC2023 avg@32 by 2.27\% and Minerva pass@32 by 3.67\%; Clip-Cov's Minerva avg@32/pass@32 falls by 1.07\%/1.83\%. Figure 5 shows all 36 contrasts, including unchanged Clip-Cov AMC2023 pass@32.

\subsection{Complete-configuration results across Qwen2.5 initializations}

Table 2 evaluates the complete Qwen2.5 configuration across model size and mathematical pretraining under the original four-benchmark protocol (full scores: Appendix D.1).

MGE-PO improves both macro metrics over GRPO across all three initializations. On Qwen2.5-7B-Base, avg@32 rises 41.5\% to 45.4\% and pass@32 58.6\% to 62.8\%, with AIME24/AIME25 gains of 9.7\%/7.0\%. Math-7B pass@32 rises 63.2\% to 70.0\%, with AIME gains of 13.3\%/7.9\%.

At fixed \(\beta=0.001\), Figure 4 shows 4Q gains of 2.1\%, 2.3\%, and 2.9\% with EA-KL, DynZPD, and both. These single-run conditional contrasts retain supporting configurations; fixed-correction and shuffled-entropy controls remain needed (Appendix D.10). Appendix D.6 tests quadrant coefficients.

Tradeoffs remain: Math-7B KL-Cov exceeds MGE-PO on macro avg@32 (43.6\% versus 43.5\%) and MATH-500 avg@32 (75.8\% versus 73.1\%). On plain 7B, AMC23 avg@32 rises from GRPO's 58.7\% to 65.5\%, while pass@32 falls from 92.5\% to 90.0\%.

MGE-PO leads five of six supplementary metrics, including Minerva/Olympiad gains of 3.1\%/3.9\% over GRPO; DAPO leads AMC23 (Appendix D.2, including Clip-Cov). The Instruct comparison improves all six metrics over GRPO (Appendix D.5).

\begin{figure}[!htb]
\centering
\includegraphics[width=\linewidth]{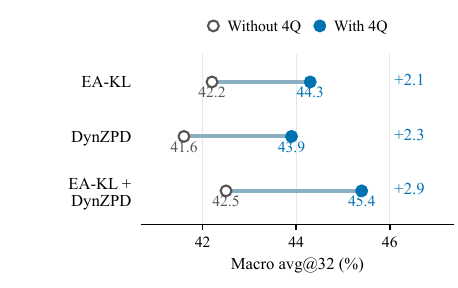}
\caption{Conditional 4Q gains on Qwen2.5-7B-Base at \(\beta=0.001\) (Table 15). Pairs share their supporting configuration; EA-KL off retains uniform KL. Scores are four-benchmark macro avg@32; annotations give percentage-point gains. Each configuration uses one training run; uncertainty intervals are unavailable.}\label{fig:4}
\end{figure}

\subsection{Outcome, training, and sampling diagnostics}

Low-entropy incorrect responses account for 45.5\% of initialization rollouts and 37.2\% after GRPO (Table 8). Concentration can coexist with failure, motivating outcome-conditioned entropy for rollout credit.

Appendix F.3 retains the original full-configuration training curves as supplementary diagnostics. Their checkpoint correspondence with Table 2 is undocumented. Appendix D.7 reports reference-weight diagnostics.

For sampling distribution \(p\), \(\operatorname{ESS}(p)=1/\sum_qp(q)^2\), with relative ESS \(100\operatorname{ESS}(p)/|\mathcal D|\). Over 21 steps and 43,008 rollouts, MGE-PO samples 2,440 unique questions: relative ESS is 91.5\%, top-decile mass 18.3\%, and coverage 14.0\% versus uniform expectation 14.3\% (Table 14). These are window-specific diagnostics; the positive floor retains dataset-wide support.

Appendix D.8 reports complete-run timing (Table 13).

\section{Conclusion}

MGE-PO uses verified outcomes and generation entropy to refine rollout credit in mathematical RLVR. Its 4Q rule gives lower-entropy successes larger additional bonuses and higher-entropy failures larger additional penalties. Under binary-reward GRPO, shared batch recentering preserves the entropy ordering of equally rewarded responses. An online entropy-based question curriculum supports this rule using cached rollout measurements.

On Qwen3-8B-Base at the final checkpoint after eight epochs, adding 4Q to GRPO, DAPO, and Clip-Cov yields absolute gains of 3.23--3.69\% in six-benchmark macro avg@32 and 4.24--11.06\% in macro pass@32. The complete configuration also improves mathematical accuracy and AIME coverage over GRPO across the evaluated Qwen2.5 base initializations. These results support outcome-conditioned entropy as a credit signal across the evaluated model and optimization settings.

\label{sec:main-end}\clearpage

\section*{Limitations}

Evaluation covers mathematical reasoning on Qwen models. The comparisons evaluate the specified 4Q rule; fixed-correction and shuffled-entropy controls remain needed to isolate entropy-dependent ordering from outcome-only shaping and arbitrary response-dependent variation. The Qwen2.5 component study uses a different global KL coefficient for GRPO, so its complete-method comparison includes that change. Shared-coefficient component contrasts provide the conditional comparisons described in Appendix D.10. The quadrant-coefficient sweep uses AIME24, also a final evaluation benchmark, without a documented independent development set.

Training-seed variability is unmeasured. The Qwen2.5 scores average repeated inference evaluations within trained checkpoints. Appendix C records the available protocol metadata and the run-specific settings that remain unspecified.

\section*{Ethical Considerations}

This study uses existing mathematical benchmarks and pretrained model checkpoints. Improved answer generation can support research and educational assistance, and can also facilitate misuse in assessments. Final-answer verification and predictive concentration do not validate every intermediate reasoning step. Applications should evaluate generated reasoning independently before presenting it as trustworthy. Reproducibility artifacts should identify the exact model and dataset releases, their licenses, and the training and evaluation configurations.

\FloatBarrier\section*{References}\begingroup\fontsize{10}{12}\selectfont\begin{list}{}{\setlength{\leftmargin}{1em}\setlength{\itemindent}{-1em}\setlength{\itemsep}{2pt}\setlength{\parsep}{0pt}}

\item Hongye Cao, Zhixin Bai, Ziyue Peng, Boyan Wang, Tianpei Yang, Jing Huo, Yuyao Zhang, and Yang Gao. 2025. \emph{Efficient Reinforcement Learning with Semantic and Token Entropy for LLM Reasoning}. arXiv:2512.04359. \href{https://arxiv.org/abs/2512.04359}{Paper}.

\item Minghan Chen, Guikun Chen, Wenguan Wang, and Yi Yang. 2025. \emph{SEED-GRPO: Semantic Entropy Enhanced GRPO for Uncertainty-Aware Policy Optimization}. arXiv:2505.12346. \href{https://arxiv.org/abs/2505.12346}{Paper}.

\item Daixuan Cheng, Shaohan Huang, Xuekai Zhu, Bo Dai, Xin Zhao, Zhenliang Zhang, and Furu Wei. 2026. \emph{Reasoning with Exploration: An Entropy Perspective}. Proceedings of the AAAI Conference on Artificial Intelligence, 40(36):30377--30385. DOI:10.1609/aaai.v40i36.40290. \href{https://ojs.aaai.org/index.php/AAAI/article/view/40290}{Paper}.

\item Ganqu Cui, Yuchen Zhang, Jiacheng Chen, Lifan Yuan, Zhi Wang, Yuxin Zuo, Haozhan Li, Yuchen Fan, Huayu Chen, Weize Chen, Zhiyuan Liu, Hao Peng, Lei Bai, Wanli Ouyang, Yu Cheng, Bowen Zhou, and Ning Ding. 2025. \emph{The Entropy Mechanism of Reinforcement Learning for Reasoning Language Models}. arXiv:2505.22617. \href{https://arxiv.org/abs/2505.22617}{Paper}.

\item Jujie He et al. 2025. \emph{Skywork Open Reasoner 1 Technical Report}. arXiv:2505.22312. \href{https://arxiv.org/abs/2505.22312}{Paper}.

\item Yuhang He, Haodong Wu, Siyi Liu, Hongyu Ge, Hange Zhou, Keyi Wu, Zhuo Zheng, Qihong Lin, Zixin Zhong, and Yongqi Zhang. 2026. \emph{Rethinking Token-Level Credit Assignment in RLVR: A Polarity-Entropy Analysis}. arXiv:2604.11056v1, April 13, 2026. \href{https://arxiv.org/abs/2604.11056v1}{Paper}.

\item Jing-Cheng Pang, Liu Sun, Chang Zhou, Xian Tang, Haichuan Ma, Kun Jiang, Jianlong Wang, Kai Zhang, Sijie Wu, Haoran Cai, Chenwei Wu, Xubin Li, and Xin Chen. 2026. \emph{EDCO: Dynamic Curriculum Orchestration for Domain-specific Large Language Model Fine-tuning}. arXiv:2601.03725. \href{https://arxiv.org/abs/2601.03725}{Paper}.

\item John Schulman, Filip Wolski, Prafulla Dhariwal, Alec Radford, and Oleg Klimov. 2017. \emph{Proximal Policy Optimization Algorithms}. arXiv:1707.06347. \href{https://arxiv.org/abs/1707.06347}{Paper}.

\item Zhihong Shao, Peiyi Wang, Qihao Zhu, Runxin Xu, Junxiao Song, Xiao Bi, Haowei Zhang, Mingchuan Zhang, Y. K. Li, Y. Wu, and Daya Guo. 2024. \emph{DeepSeekMath: Pushing the Limits of Mathematical Reasoning in Open Language Models}. arXiv:2402.03300. \href{https://arxiv.org/abs/2402.03300}{Paper}.

\item Qiying Yu, Zheng Zhang, Ruofei Zhu, Yufeng Yuan, Xiaochen Zuo, Yu Yue, Weinan Dai, Tiantian Fan, Gaohong Liu, Lingjun Liu, Xin Liu, Haibin Lin, Zhiqi Lin, Bole Ma, Guangming Sheng, Yuxuan Tong, Chi Zhang, Mofan Zhang, Wang Zhang, Hang Zhu, Jinhua Zhu, Jiaze Chen, Jiangjie Chen, Chengyi Wang, Hongli Yu, Yuxuan Song, Xiangpeng Wei, Hao Zhou, Jingjing Liu, Wei-Ying Ma, Ya-Qin Zhang, Lin Yan, Mu Qiao, Yonghui Wu, and Mingxuan Wang. 2025. \emph{DAPO: An Open-Source LLM Reinforcement Learning System at Scale}. arXiv:2503.14476. \href{https://arxiv.org/abs/2503.14476}{Paper}.

\item Matthieu Geist, Bruno Scherrer, and Olivier Pietquin. 2019. \emph{A Theory of Regularized Markov Decision Processes}. Proceedings of the 36th International Conference on Machine Learning, PMLR 97:2160--2169. \href{https://proceedings.mlr.press/v97/geist19a.html}{Paper}.

\item Youssef Mroueh. 2026. \emph{Reinforcement Learning with Verifiable Rewards: GRPO’s Loss, Dynamics, and Success Amplification}. Proceedings of the 43rd International Conference on Machine Learning, PMLR 306:90821--90847. \href{https://proceedings.mlr.press/v306/mroueh26a.html}{Paper}.

\item Jean Vassoyan, Nathanaël Beau, and Roman Plaud. 2025. \emph{Ignore the KL Penalty! Boosting Exploration on Critical Tokens to Enhance RL Fine-Tuning}. Findings of the Association for Computational Linguistics: NAACL 2025, pages 6123--6133. \href{https://aclanthology.org/2025.findings-naacl.340/}{Paper}.

\item Daya Guo et al. 2025. \emph{DeepSeek-R1 incentivizes reasoning in LLMs through reinforcement learning}. Nature, 645:633--638. \href{https://doi.org/10.1038/s41586-025-09422-z}{Paper}.

\item Dan Hendrycks, Collin Burns, Saurav Kadavath, Akul Arora, Steven Basart, Eric Tang, Dawn Song, and Jacob Steinhardt. 2021. \emph{Measuring Mathematical Problem Solving With the MATH Dataset}. Proceedings of the Neural Information Processing Systems Track on Datasets and Benchmarks. \href{https://arxiv.org/abs/2103.03874}{Paper}.

\item Aitor Lewkowycz et al. 2022. \emph{Solving Quantitative Reasoning Problems with Language Models}. Advances in Neural Information Processing Systems. \href{https://arxiv.org/abs/2206.14858}{Paper}.

\item An Yang et al. 2024a. \emph{Qwen2.5 Technical Report}. arXiv:2412.15115, first submitted December 19, 2024. \href{https://arxiv.org/abs/2412.15115v1}{Paper}.

\item An Yang et al. 2024b. \emph{Qwen2.5-Math Technical Report: Toward Mathematical Expert Model via Self-Improvement}. arXiv:2409.12122. \href{https://arxiv.org/abs/2409.12122}{Paper}.

\item An Yang et al. 2025. \emph{Qwen3 Technical Report}. arXiv:2505.09388. \href{https://arxiv.org/abs/2505.09388}{Paper}.

\item Mark Chen et al. 2021. \emph{Evaluating Large Language Models Trained on Code}. arXiv:2107.03374. \href{https://arxiv.org/abs/2107.03374}{Paper}.

\item Andrey Malinin and Mark Gales. 2021. \emph{Uncertainty Estimation in Autoregressive Structured Prediction}. International Conference on Learning Representations. \href{https://arxiv.org/abs/2002.07650}{Paper}.

\item Lorenz Kuhn, Yarin Gal, and Sebastian Farquhar. 2023. \emph{Semantic Uncertainty: Linguistic Invariances for Uncertainty Estimation in Natural Language Generation}. International Conference on Learning Representations. \href{https://arxiv.org/abs/2302.09664}{Paper}.

\item Yikai Zhou, Baosong Yang, Derek F. Wong, Yu Wan, and Lidia S. Chao. 2020. \emph{Uncertainty-Aware Curriculum Learning for Neural Machine Translation}. Proceedings of the 58th Annual Meeting of the Association for Computational Linguistics, pages 6934--6944. \href{https://aclanthology.org/2020.acl-main.620/}{Paper}.

\item Emmanouil Antonios Platanios, Otilia Stretcu, Graham Neubig, Barnabas Poczos, and Tom Mitchell. 2019. \emph{Competence-based Curriculum Learning for Neural Machine Translation}. Proceedings of NAACL-HLT, Volume 1, pages 1162--1172. \href{https://aclanthology.org/N19-1119/}{Paper}.

\item Hunter Lightman et al. 2024. \emph{Let’s Verify Step by Step}. International Conference on Learning Representations. \href{https://proceedings.iclr.cc/paper_files/paper/2024/hash/aca97732e30bcf1303bc22ac3924fd16-Abstract-Conference.html}{Paper}.

\item Peiyi Wang, Lei Li, Zhihong Shao, Runxin Xu, Damai Dai, Yifei Li, Deli Chen, Yu Wu, and Zhifang Sui. 2024. \emph{Math-Shepherd: Verify and Reinforce LLMs Step-by-step without Human Annotations}. Proceedings of the 62nd Annual Meeting of the Association for Computational Linguistics, Volume 1: Long Papers, pages 9426--9439. \href{https://aclanthology.org/2024.acl-long.510/}{Paper}.

\end{list}\endgroup

\clearpage\onecolumn\raggedbottom\appendix

\section{Proofs and Additional Optimization Interpretation}

Section 5 characterizes 4Q's credit coefficients directly. We first prove the sampling proposition, then provide a standard reference-constrained interpretation of a frozen outcome-conditioned utility. The exact-optimum results assume a finite completion space, finite fixed utilities, a full-support reference distribution, and fixed \(\beta>0\). They explain preference and deviation cost under exact distributional optimization; the empirical actor uses sampled, clipped updates with shared neural parameters.

\Needspace{120pt}

\subsection{Mixed-reward opportunities and finite-budget allocation}

\textbf{Proof of Proposition 1.} Independence gives probability \(s^G\) for an all-success group and \((1-s)^G\) for an all-failure group. These events are disjoint, so \(m_G(s)=1-s^G-(1-s)^G\). For \(s\in(0,1)\),

\[
\adjustbox{max width=\dimexpr\linewidth-24pt\relax}{$\displaystyle m_G'(s)=G\left[(1-s)^{G-1}-s^{G-1}\right].$}
\]

For \(G\ge2\), this derivative is positive below \(1/2\), zero at \(1/2\), and negative above \(1/2\). Thus \(1/2\) is the unique maximizer, with value \(1-2^{1-G}\); both endpoint values are zero. Mixed groups supply nonzero base GRPO coefficients, although their aggregate gradient depends on the sampled actions and model parameters.

Sampling frequency matters for a policy whose parameters are shared across questions. Let \(V(\theta)\) be a locally smooth held-out objective and \(g_q\) an ascent update estimated from question \(q\), including the learning signal and regularization. For a sufficiently small step \(\lambda_{\mathrm{step}}\),

\[
\adjustbox{max width=\dimexpr\linewidth-24pt\relax}{$\displaystyle V(\theta+\lambda_{\mathrm{step}}g_q)
=V(\theta)+\lambda_{\mathrm{step}}\nabla V(\theta)^\top g_q
+O(\lambda_{\mathrm{step}}^2\|g_q\|^2).$}
\]

Averaging over update noise and question draws gives first-order contribution proportional to

\[
\adjustbox{max width=\dimexpr\linewidth-24pt\relax}{$\displaystyle \sum_qp_k(q)\nabla V(\theta)^\top\mathbb E[g_q].$}
\]

This expression describes immediate learning value through gradient alignment. DynZPD uses online entropy as an empirical proxy for useful training regimes; neither the alignment expression nor Proposition 1 establishes an optimal entropy target. For the independent distributional model in Equation (A3), let \(J_q(\pi_q)\) denote its per-question objective. Positive question weights in \(\sum_qp_k(q)J_q(\pi_q)\) do not change the per-question exact optima. A shared neural policy and a finite update budget make allocation consequential. The weight floor preserves positive sampling support; realized coverage depends on the finite sequence of draws.

\Needspace{120pt}

\subsection{Frozen-utility preferences and outcome-sign consistency}

For a question \(q\) and completion \(y\), let \(f_{0,k}(q,y)\) be a fixed baseline utility, and freeze the entropy and truncation signals. Retain \(r(q,y)\in\{0,1\}\), \(u_k(q,y)\in[0,1]\), \(g(q,y)\in\{0,1\}\), and \(\gamma_+,\gamma_-\ge0\). Define

\begin{equation}
\adjustbox{max width=\dimexpr\linewidth-36pt\relax}{$\displaystyle \delta_k(q,y)=g(q,y)\left[
r(q,y)\gamma_+u_k(q,y)
-(1-r(q,y))\gamma_-(1-u_k(q,y))
\right],$}
\tag{A1}\end{equation}

\begin{equation}
\adjustbox{max width=\dimexpr\linewidth-36pt\relax}{$\displaystyle f_k(q,y)=f_{0,k}(q,y)+\delta_k(q,y).$}
\tag{A2}\end{equation}

Let \(\nu_q(y)=\pi_{\mathrm{ref}}(y\mid q)>0\). At fixed \(\beta>0\), consider

\begin{equation}
\adjustbox{max width=\dimexpr\linewidth-36pt\relax}{$\displaystyle \max_{\pi(\cdot\mid q)}
\left\{
\mathbb E_{y\sim\pi(\cdot\mid q)}[f_k(q,y)]
-\beta D_{\mathrm{KL}}\bigl(\pi(\cdot\mid q)\Vert\nu_q\bigr)
\right\}.$}
\tag{A3}\end{equation}

\textbf{Proposition A.1 (Regularized trajectory preference).} The unique maximizer is

\begin{equation}
\adjustbox{max width=\dimexpr\linewidth-36pt\relax}{$\displaystyle \begin{gathered}
\pi_k^*(y\mid q)=
\frac{\nu_q(y)\exp(f_k(q,y)/\beta)}{Z_k(q)},\\
Z_k(q)=\sum_y\nu_q(y)\exp(f_k(q,y)/\beta).
\end{gathered}$}
\tag{A4}\end{equation}

This standard KL-regularized solution (Geist et al., 2019; Mroueh, 2026) implies

\begin{equation}
\adjustbox{max width=\dimexpr\linewidth-36pt\relax}{$\displaystyle \log\frac{\pi_k^*(y_1\mid q)}{\pi_k^*(y_2\mid q)}
=\log\frac{\nu_q(y_1)}{\nu_q(y_2)}
+\frac{f_k(q,y_1)-f_k(q,y_2)}{\beta}.$}
\tag{A5}\end{equation}

\textbf{Proof.} Suppressing \(q,k\), substitute \(\pi^*(y)=\nu(y)e^{f(y)/\beta}/Z\) to obtain

\[
\adjustbox{max width=\dimexpr\linewidth-24pt\relax}{$\displaystyle \mathbb E_\pi[f]-\beta D_{\mathrm{KL}}(\pi\Vert\nu)
=\beta\log Z-\beta D_{\mathrm{KL}}(\pi\Vert\pi^*).$}
\]

KL nonnegativity gives the unique maximum at \(\pi^*\). The stated assumptions ensure that \(Z\) is finite and positive and \(\pi^*\) has full support. Subtracting a common frozen constant from \(f\) multiplies both numerator and normalizer by the same factor, leaving the optimum unchanged. This includes batch recentering after its common shift is frozen.

\textbf{Proposition A.2 (Outcome-sign consistency at an exact optimum).} Holding the reference, baseline utility, and positive \(\beta\) fixed, adding a correction that is nonnegative on successes and nonpositive on failures weakly increases total success probability at the exact optimum.

\textbf{Proof.} Let \(\mathcal C=\{y:r(q,y)=1\}\) and \(\mathcal I=\{y:r(q,y)=0\}\). If either set is empty, success probability is fixed at zero or one. Otherwise both baseline masses

\[
\adjustbox{max width=\dimexpr\linewidth-24pt\relax}{$\displaystyle M_C=\sum_{y\in\mathcal C}\nu(y)e^{f_0(y)/\beta},
\qquad
M_I=\sum_{y\in\mathcal I}\nu(y)e^{f_0(y)/\beta}$}
\]

are positive. The shaped masses satisfy

\[
\adjustbox{max width=\dimexpr\linewidth-24pt\relax}{$\displaystyle \begin{aligned}
M_C'&=\sum_{y\in\mathcal C}\nu(y)e^{f_0(y)/\beta}e^{\delta(y)/\beta}\ge M_C,\\
M_I'&=\sum_{y\in\mathcal I}\nu(y)e^{f_0(y)/\beta}e^{\delta(y)/\beta}\le M_I.
\end{aligned}$}
\]

Consequently \(M_C'/M_I'\ge M_C/M_I\), and monotonicity of \(x/(1+x)\) gives the claim. The inequality is strict if a success has \(\delta>0\) or a failure has \(\delta<0\). Equation (A1) satisfies these signs before recentering; a common frozen shift cancels by Proposition A.1. This comparison supports the outcome-conditioned signs, while several magnitude rules satisfy the same condition. The entropy ordering and the sampled update's performance remain empirical questions.

\Needspace{120pt}

\subsection{Reference costs and the sampled actor objective}

At a fixed prefix \(c\), let \(Q_c(v)\) be a fixed finite next-token utility, \(\nu_c(v)>0\) the reference distribution, and \(w_k(c)>0\) a detached weight. The local problem

\begin{equation}
\adjustbox{max width=\dimexpr\linewidth-36pt\relax}{$\displaystyle \max_{p\in\Delta(\mathcal V)}
\left\{\sum_vp(v)Q_c(v)
-\beta w_k(c)D_{\mathrm{KL}}(p\Vert\nu_c)\right\}$}
\tag{A6}\end{equation}

has the solution

\begin{equation}
\adjustbox{max width=\dimexpr\linewidth-36pt\relax}{$\displaystyle p_c^*(v)=
\frac{\nu_c(v)\exp\!\left(Q_c(v)/(\beta w_k(c))\right)}
{\sum_{v'}\nu_c(v')\exp\!\left(Q_c(v')/(\beta w_k(c))\right)}.$}
\tag{A7}\end{equation}

This follows from Proposition A.1 with temperature \(\beta w_k(c)\). In particular,

\[
\adjustbox{max width=\dimexpr\linewidth-24pt\relax}{$\displaystyle \log\frac{p_c^*(v_1)}{p_c^*(v_2)}
-\log\frac{\nu_c(v_1)}{\nu_c(v_2)}
=\frac{Q_c(v_1)-Q_c(v_2)}{\beta w_k(c)}.$}
\]

Increasing the weight reduces the utility-induced shift in reference-relative log odds. It need not increase Shannon entropy, which also depends on the reference and utility ordering. At \(w_k(c)=0\), the local objective is linear and any distribution supported on utility maximizers is optimal. As positive \(w\) approaches zero, the Gibbs solution concentrates on these maximizers with their relative reference weights.

For consistently treated termination, full-sequence KL obeys the autoregressive chain rule

\[
\adjustbox{max width=\dimexpr\linewidth-24pt\relax}{$\displaystyle \begin{aligned}
&D_{\mathrm{KL}}(\pi(\cdot\mid q)\Vert\pi_{\mathrm{ref}}(\cdot\mid q))\\
&\quad=\mathbb E_{y\sim\pi(\cdot\mid q)}
\left[\sum_{t=1}^{|y|}
D_{\mathrm{KL}}\bigl(\pi(\cdot\mid q,y_{<t})
\Vert\pi_{\mathrm{ref}}(\cdot\mid q,y_{<t})\bigr)\right].
\end{aligned}$}
\]

This identity uses candidate-policy prefix occupancy and a sum over sequence positions. The actor evaluates rollout-policy samples with masks, normalization, entropy weights, and a clipped likelihood-ratio surrogate. Appendix B.4 gives the implemented token proxy and its sampling convention. The exact-optimum statements describe frozen preferences and deviation costs; the component experiments assess the sampled neural update. Uniform-weight controls and entropy-weighted controls retain their stated global coefficients, while complete KL removal changes the reference constraint.

\section{Implementation Details}

\Needspace{120pt}

\subsection{Running rollout-entropy statistics}

For an actor batch with \(M\) rollouts, define

\[
\adjustbox{max width=\dimexpr\linewidth-24pt\relax}{$\displaystyle \begin{gathered}
\bar h_b=\frac{1}{M}\sum_{i=1}^{M}h_i^{\mathrm{top}},
\\
t_b=\sqrt{\frac{1}{M-1}\sum_{i=1}^{M}(h_i^{\mathrm{top}}-\bar h_b)^2}.
\end{gathered}$}
\]

The running statistics are updated as

\[
\adjustbox{max width=\dimexpr\linewidth-24pt\relax}{$\displaystyle \begin{gathered}
\mu_H\leftarrow\lambda_H\mu_H+(1-\lambda_H)\bar h_b,
\\
\tau_H\leftarrow\lambda_H\tau_H+(1-\lambda_H)t_b.
\end{gathered}$}
\]

Both statistics are initialized from the first valid batch. The reported momentum is \(\lambda_H=0.99\) and the denominator stabilizer is \(\epsilon_u=10^{-6}\). The running scale is an EMA of batch standard deviations. Algorithm 1 updates these statistics before computing \(u_i\).

\Needspace{120pt}

\subsection{Generalized continuous four-quadrant rule}

For sensitivity comparisons, define

\[
\adjustbox{max width=\dimexpr\linewidth-24pt\relax}{$\displaystyle \delta_i^{\mathrm{general}}
=g_i\left\{
r_i\left[\gamma_{\mathrm{PLR}}u_i+\gamma_{\mathrm{PHR}}(1-u_i)\right]
-(1-r_i)\left[\gamma_{\mathrm{NHR}}(1-u_i)+\gamma_{\mathrm{NLR}}u_i\right]
\right\}.$}
\]

The historical default sets \((\gamma_{\mathrm{PLR}},\gamma_{\mathrm{NHR}},\gamma_{\mathrm{PHR}},\gamma_{\mathrm{NLR}})=(0.5,0.3,0,0)\). The named terms form a continuous interpolation within each outcome class. The \(u_i=0.5\) threshold defines the diagnostic count bins.

\Needspace{120pt}

\subsection{Degenerate groups, truncation, and recentering}

If all outcomes in a group are identical, the base GRPO advantages are zero. Entropy corrections may still vary across rollouts, so the modified advantages need not be zero. After batch recentering, an incorrect rollout can receive a positive effective advantage relative to the batch, even when its direct correction is nonpositive. Effective advantages therefore need to be interpreted after centering.

For example, in an all-incorrect, untruncated batch, the default direct correction is \(-\gamma_-(1-u_i)\). Recentering gives

\[
\adjustbox{max width=\dimexpr\linewidth-24pt\relax}{$\displaystyle \widetilde A_i=\gamma_-\left(u_i-\frac{1}{M}\sum_j u_j\right).$}
\]

A rollout with above-average \(u_i\) can therefore receive positive relative credit without being correct. The example concerns a finite empirical batch, while Proposition A.2 compares exact optima with a frozen baseline. Effective-advantage diagnostics can quantify this behavior during training.

The gate \(g_i=1-\mathbf 1[y_i\text{ is truncated}]\) suppresses only the direct entropy correction. The gate tests truncation; zero-variance groups follow the shaping and centering rules above. Grading and normalization conventions determine which truncated rollouts participate in the batch. Padding masks exclude invalid token positions from actor and KL reductions.

\Needspace{120pt}

\subsection{Implemented entropy-weighted reference regularization}

EA-KL varies the token-level price of reference deviation according to cached entropy. With active-token mask \(m_{i,t}\), define a stabilized mean and the corresponding weights:

\[
\adjustbox{max width=\dimexpr\linewidth-24pt\relax}{$\displaystyle \bar H
=\max\left(
\frac{\sum_{i,t}m_{i,t}H_{i,t}}{\sum_{i,t}m_{i,t}},
H_{\min}
\right),$}
\]

\[
\adjustbox{max width=\dimexpr\linewidth-24pt\relax}{$\displaystyle w_{i,t}
=\operatorname{clip}\left[
1+\alpha\left(1-\frac{H_{i,t}}{\bar H}\right),
0,w_{\max}
\right].$}
\]

For \(\alpha\ge0\), tokens below \(\bar H\) receive a larger relative deviation cost and those above it a smaller one. Uniform weights are recovered at \(\alpha=0\). The reported configuration uses \(\alpha=0.20\), \(H_{\min}=5\times10^{-3}\), and \(w_{\max}=1.5\); the realized upper bound and zero-weight cases are described below.

The actor penalty uses the sampled nonnegative proxy

\[
\adjustbox{max width=\dimexpr\linewidth-24pt\relax}{$\displaystyle \begin{gathered}
k_{i,t}(\theta)
=\exp(d_{i,t})-d_{i,t}-1,
\\
d_{i,t}
=\log\pi_{\mathrm{ref}}(a_{i,t}\mid c_{i,t})
-\log\pi_\theta(a_{i,t}\mid c_{i,t}),
\end{gathered}$}
\]

aggregated as

\[
\adjustbox{max width=\dimexpr\linewidth-24pt\relax}{$\displaystyle \widehat{\mathcal K}_w
=\frac{\sum_{i,t}m_{i,t}w_{i,t}k_{i,t}(\theta)}
{\sum_{i,t}m_{i,t}}.$}
\]

This proxy equals forward KL in expectation when the action is sampled from the current policy. The implemented actor update evaluates rollout-policy samples, and Appendix A.3 distinguishes this sampled penalty from its exact distributional model. Gradient treatment, reduction, and any importance correction are part of the configuration metadata in Appendix C.

Weights and entropy statistics are detached. For the historical \(\alpha=0.20\), the raw expression satisfies \(w_{i,t}\le1.20\) for nonnegative entropy before upper clipping. The lower clipping threshold is reached when \(H_{i,t}/\bar H\ge6\). Appendix D.7 supplies the PLR-stratified weight summaries. Overall mean weight and zero-weight fraction are unreported.

Adaptive weights can correlate with token KL values, so their arithmetic mean alone does not determine aggregate penalty strength.

\Needspace{120pt}

\subsection{Curriculum normalization and support}

For a finite dataset with \(N_D\) questions, \(v_q\in[\eta,1]\) implies

\[
\adjustbox{max width=\dimexpr\linewidth-24pt\relax}{$\displaystyle \begin{gathered}
p_k(q)>0,
\\
p_k(q)\ge\frac{\eta}{N_D}.
\end{gathered}$}
\]

This bound preserves support at each draw. The normalized distribution can be expressed as a mixture with uniform sampling, whose mixture coefficient depends on the sum of Gaussian weights. Finite-run coverage depends on the realized draws.

Stored statistics are refreshed when their questions are sampled; their update ages are not recorded in the reported diagnostics.

\Needspace{120pt}

\subsection{Update procedure}

\textbf{Algorithm 1: GRPO-based MGE-PO.} This procedure specifies the complete GRPO-based configuration. Each rollout round freezes the policy used for generation and its entropy measurements during the actor update. Question-statistic updates use the recorded rollout entropies and affect subsequent draws.

\begin{Verbatim}[fontsize=\small]
Input: training questions D, initial policy pi, fixed reference pi_ref,
       G completions per question, actor-update budget,
       shaping, KL-weight, and curriculum parameters.
Initialize xi_q = mu_ZPD for every question q in D.
For each rollout round k:
    1. Normalize stored question weights to obtain p_k; sample questions.
    2. Generate G completions per question with rollout policy pi_k.
       Cache valid-token entropy, rollout log probabilities, and masks.
    3. Grade final answers and compute group-relative base advantages.
    4. Compute top-entropy summaries h_top and full-token means h_mean.
       On the first valid batch, initialize mu_H and tau_H from its
       top-entropy statistics; on later batches, update their EMAs.
    5. Compute detached concentration scores u_i and gated corrections.
       Recenter the adjusted advantages over the actor batch.
    6. Compute detached token KL weights from cached entropy.
    7. Optimize the clipped actor surrogate minus beta times the
       weighted reference-deviation proxy; keep cached signals fixed.
    8. Update xi_q for sampled questions using their rollout mean entropy.
Return: final policy and training/evaluation configuration metadata.
\end{Verbatim}

\section{Reproducibility Configuration}

\Needspace{120pt}

\subsection{Reported settings}

\begingroup
\fontsize{10}{12}\selectfont
\setlength{\tabcolsep}{4pt}
\renewcommand{\arraystretch}{1.15}
\setlength{\tablecontentwidth}{\dimexpr\linewidth-16pt\relax}
\begin{longtable}{@{}L{0.14000\tablecontentwidth}L{0.33000\tablecontentwidth}L{0.53000\tablecontentwidth}@{}}
\caption{Reported hyperparameters and configuration metadata. Values describe the original Qwen2.5 experiments and their supplementary settings. Appendix C.4 describes the separate Qwen3-8B-Base training and evaluation protocol.}\label{tab:3}\\
\toprule
\textbf{Category} & \textbf{Quantity} & \textbf{Reported value or status} \\
\midrule
\endfirsthead
\multicolumn{3}{l}{\small\itshape Table 3 (continued)}\\
\toprule
\textbf{Category} & \textbf{Quantity} & \textbf{Reported value or status} \\
\midrule
\endhead
\midrule
\multicolumn{3}{r}{\small Continued on next page}\\
\endfoot
\bottomrule
\endlastfoot
Training & Reported global batch size & 128; question/response unit unspecified \\
Training & Completions per question & 16 \\
Training & Dataset-equivalent epochs & 8 \\
Training & Sampling temperature & 1.0 \\
Training & Maximum generation length & 12,288 tokens \\
Training & Top-\(p\) during rollout & Not reported \\
Training & Exact actor/optimizer update schedule & Not reported \\
Training & Actor-loss reduction & Not reported \\
Training & Optimizer, learning rate, scheduler & Not reported \\
Training & Checkpoint identifiers & Supplementary Instruct: final step 1080; supplementary Base step unreported \\
EA-Adv & \(\gamma_+\) & 0.5 \\
EA-Adv & \(\gamma_-\) & 0.3 \\
EA-Adv & Separate PHR/NLR coefficients & 0, 0 \\
EA-Adv & \(r_K\), \(K_{\max}\) & 0.1, 4096 \\
EA-Adv & \(\lambda_H\), \(\epsilon_u\) & 0.99, \(10^{-6}\) \\
EA-Adv & Recentering and masks & Batch recentering specified; exact implementation masks unreported \\
Entropy & Probability distribution & Policy distribution in the formulation; decoding-transform convention unreported \\
EA-KL & \(\alpha\) & 0.20 \\
EA-KL & \(H_{\min}\) & \(5\times10^{-3}\) \\
EA-KL & Configured weight bounds & 0, 1.5 \\
KL & Original MGE-PO coefficient & \(10^{-3}\) \\
KL & Original GRPO coefficient & 0.04 \\
KL & Seven nonempty component configurations & \(10^{-3}\); uniform weights when EA-KL is off \\
DynZPD & \(\mu_{\mathrm{ZPD}}\), \(\sigma_{\mathrm{ZPD}}\) & 0.30, 0.15 \\
DynZPD & Weight floor \(\eta\) & 0.10 \\
DynZPD & Question EMA \(\lambda_q\) & 0.9 \\
DynZPD & Initial question statistic & \(\mu_{\mathrm{ZPD}}\) \\
Evaluation & Original temperature, top-\(p\) & 1.0, 0.7; supplementary values unspecified \\
Evaluation & Supplementary Minerva/Olympiad maximum length & 8,192 tokens \\
Evaluation & Supplementary accuracy samples & 64 per question \\
Evaluation & Supplementary AIME coverage & pass@256, averaged across repeated inference evaluations \\
Evaluation & Pass@K repeat counts and inference seeds & Not reported for either protocol \\
Compute & Historical training hardware & \(2\times\) NVIDIA H200, 141 GB each \\
Statistics & Training runs per RL method cell & One \\
\end{longtable}
\endgroup

The entropy equation in Section 3.2 defines the policy distribution before decoding transformations. Reproducing cached entropy measurements additionally requires the extraction stage relative to temperature and sampling transformations, together with the active-token masks. The run-specific decoding-transform convention is not specified in the current configuration record.

\Needspace{120pt}

\subsection{Dataset and evaluation metadata}

The original Qwen2.5 evaluation uses MATH-500, AMC23, and AIME 2024/2025; its supplementary evaluation adds Minerva-Math/OCW and a text-only English exact-answer Olympiad subset. Benchmark names and counts are reported. Exact question manifests, model revisions, prompt templates, grader versions, extraction rules, and training--test overlap checks are not documented in the current record.

In the Qwen2.5 series, pass@\(K\) averages repeated inference evaluations of a trained checkpoint. The repeat counts and inference seeds are unreported. The supplementary base comparison omits its checkpoint step. The Instruct comparison identifies final step 1080 after eight epochs and describes a ``4k'' setting without specifying its role.

\Needspace{120pt}

\subsection{Measurement scope}

Each original Qwen2.5 RL model--method cell uses one training run. Its reported macro scores retain the original rounding, and pass@\(K\) repeats are inference repetitions. The Qwen3-8B-Base table reports one point estimate per method and metric at the final controller step 624; its inference-repeat count and number of training seeds are unspecified. Training-seed standard deviations and paired problem-level intervals are unavailable.

Runtime values refer to complete single runs with matched training budgets. Peak-memory and detailed GPU-hour accounting are unavailable. Curriculum statistics cover a 21-step window containing 43,008 rollouts and 2,440 observed questions; full-run coverage and stored-statistic ages are unreported. The reported maximum question share of 0.4\% has no documented denominator. For the default sampler with \(N_D=17{,}398\) and \(\eta=0.10\), \(v_q\in[\eta,1]\) implies \(p_k(q)\le1/[1+(N_D-1)\eta]\approx0.0574\%\). The 0.4\% entry therefore cannot be interpreted as the maximum single-question probability under this distribution; its measurement definition remains unresolved.

\Needspace{120pt}

\subsection{Qwen3-8B-Base training and evaluation protocol}

\begingroup
\fontsize{10}{12}\selectfont
\setlength{\tabcolsep}{4pt}
\renewcommand{\arraystretch}{1.15}
\setlength{\tablecontentwidth}{\dimexpr\linewidth-8pt\relax}
\begin{longtable}{@{}L{0.37000\tablecontentwidth}L{0.63000\tablecontentwidth}@{}}
\caption{Training and evaluation protocol for the Qwen3-8B-Base experiments. The seed below controls dataset subsampling. All seven configurations share the learning rate, evaluation question sets, and decoding settings, and are evaluated at the final controller step 624.}\label{tab:4}\\
\toprule
\textbf{Configuration} & \textbf{Value} \\
\midrule
\endfirsthead
\multicolumn{2}{l}{\small\itshape Table 4 (continued)}\\
\toprule
\textbf{Configuration} & \textbf{Value} \\
\midrule
\endhead
\midrule
\multicolumn{2}{r}{\small Continued on next page}\\
\endfoot
\bottomrule
\endlastfoot
Base model & Qwen3-8B-Base \\
Training data & Random 10,000-question subset of the Skywork-OR1 mathematical training set \\
Data-subsampling seed & 1 \\
Training questions after filtering & 9,998 \\
Train batch size & 128 questions \\
Training rollouts per question & 8 \\
Maximum training response length & 20,480 tokens \\
Training epochs & 8 \\
Complete training schedule & 624 controller steps \\
Periodic checkpoint interval & Every 50 controller steps \\
Learning rate & Same across all seven configurations \\
Evaluation benchmarks & AIME2024, AIME2025, AMC2023, MATH500, Minerva, OlympiadBench; identical question sets across configurations \\
Evaluation responses per question & 32 \\
Evaluation temperature & 0.6 \\
Evaluation decoding settings & Same across all seven configurations \\
Evaluation metrics & mean@32 (denoted avg@32 in this paper), pass@32 \\
Macro aggregation & Equal-weight mean over the six benchmarks \\
\end{longtable}
\endgroup

The subset is sampled before filtering, leaving 9,998 training questions. The reported 624 controller steps constitute the full eight-epoch schedule. Table 1 uses the final checkpoint for every configuration; the periodic checkpoint interval is 50 controller steps. Data-subsampling seed 1 does not specify the model-training or inference seeds.

Evaluation uses identical test question sets and decoding settings across all seven configurations, sampling 32 responses per question at temperature 0.6. The reported mean@32 is the sample-average accuracy defined as avg@32 in Section 6.1; pass@32 measures question coverage within the 32 responses. The six benchmarks receive equal weight in each macro score. The 20,480-token limit in Table 4 applies to training responses; the maximum evaluation response length, numerical learning rate, and remaining decoding parameter values are unspecified.

Table 1 retains the per-benchmark scores and macro averages at their reported precision. The manuscript displays two decimal places; Table 16 computes differences before rounding the macro averages to that precision. Small differences between reported macro scores and means recomputed from the displayed per-benchmark values reflect their different display precision. The 4Q comparisons are GRPO versus 4Q, DAPO versus 4Q + DAPO, and Clip-Cov versus Clip-Cov + 4Q. These final-checkpoint comparisons do not estimate training-seed uncertainty.

Qwen3 inherits the Qwen2.5 component defaults in Table 3, except for the training and evaluation settings in Table 4. For 4Q, \(\gamma_+=0.5\) and \(\gamma_-=0.3\), with separate PHR/NLR coefficients set to zero. Entropy pooling uses \(r_K=0.1\), \(K_{\max}=4096\), \(\lambda_H=0.99\), and \(\epsilon_u=10^{-6}\) (Appendix B.1). Where DynZPD is active, it uses the question statistic and update rule in Section 4.3 with \(\mu_{\mathrm{ZPD}}=0.30\), \(\sigma_{\mathrm{ZPD}}=0.15\), \(\eta=0.10\), and \(\lambda_q=0.9\). EA-KL uses \(\alpha=0.20\), \(H_{\min}=5\times10^{-3}\), and weight bounds \([0,1.5]\); the complete MGE-PO configuration retains \(\beta=0.001\).

Component-default inheritance does not specify each host optimizer's KL setting or its integration details. For the 4Q-host combinations, the run-specific host reward and base-advantage computation, correction insertion point relative to normalization, recentering population, truncation and token masks, and any token-selection statistic derived from the adjusted advantage remain undocumented. Section 4.2 separates binary verification from the host's base reward; Section 5.2 retains its binary-reward GRPO conditions. These remaining record gaps limit exact reproduction of the host integrations.

\section{Complete Reported Results and Additional Comparisons}

\Needspace{120pt}

\subsection{Original per-benchmark scores}

\begingroup
\fontsize{10}{12}\selectfont
\setlength{\tabcolsep}{1.5pt}
\renewcommand{\arraystretch}{1.15}
\setlength{\tablecontentwidth}{\dimexpr\linewidth-33.0pt\relax}
\begin{longtable}{@{}L{0.22000\tablecontentwidth}L{0.09500\tablecontentwidth}C{0.06850\tablecontentwidth}C{0.06850\tablecontentwidth}C{0.06850\tablecontentwidth}C{0.06850\tablecontentwidth}C{0.06850\tablecontentwidth}C{0.06850\tablecontentwidth}C{0.06850\tablecontentwidth}C{0.06850\tablecontentwidth}C{0.06850\tablecontentwidth}C{0.06850\tablecontentwidth}@{}}
\caption{Full original results under avg@32/pass@32. Each RL method cell is based on one training run; Base rows evaluate the pre-RL initialization. Pass@32 averages repeated inference evaluations of that checkpoint. Macro averages are retained as reported; rounding of displayed per-benchmark values can produce small differences from a recomputed mean. A25/A24 denote AIME 2025/2024, AMC denotes AMC23, and M500 denotes MATH-500; avg/pass denote avg@32/pass@32. All scores use \%.}\label{tab:5}\\
\toprule
\textbf{Initialization} & \textbf{Method} & \textbf{A25\newline  avg (\%)} & \textbf{A25\newline  pass (\%)} & \textbf{A24\newline  avg (\%)} & \textbf{A24\newline  pass (\%)} & \textbf{AMC\newline  avg (\%)} & \textbf{AMC\newline  pass (\%)} & \textbf{M500\newline  avg (\%)} & \textbf{M500\newline  pass (\%)} & \textbf{Macro\newline  avg (\%)} & \textbf{Macro\newline  pass (\%)} \\
\midrule
\endfirsthead
\multicolumn{12}{l}{\small\itshape Table 5 (continued)}\\
\toprule
\textbf{Initialization} & \textbf{Method} & \textbf{A25\newline  avg (\%)} & \textbf{A25\newline  pass (\%)} & \textbf{A24\newline  avg (\%)} & \textbf{A24\newline  pass (\%)} & \textbf{AMC\newline  avg (\%)} & \textbf{AMC\newline  pass (\%)} & \textbf{M500\newline  avg (\%)} & \textbf{M500\newline  pass (\%)} & \textbf{Macro\newline  avg (\%)} & \textbf{Macro\newline  pass (\%)} \\
\midrule
\endhead
\midrule
\multicolumn{12}{r}{\small Continued on next page}\\
\endfoot
\bottomrule
\endlastfoot
Qwen2.5-Math-1.5B\newline  Base & Base & 3.3 & 28.7 & 6.9 & 10.3 & 32.5 & 72.5 & 32.8 & 77.8 & 18.9 & 47.3 \\
 & GRPO & 6.3 & 30.7 & 9.5 & 30.3 & 52.7 & 90.0 & 66.5 & 89.4 & 33.7 & 60.1 \\
 & DAPO & 7.1 & 26.7 & 11.0 & 32.7 & 54.5 & 87.5 & 68.7 & 90.0 & 35.3 & 59.2 \\
 & KL-Cov & 8.3 & 35.7 & 10.9 & 31.8 & 55.1 & 87.5 & 68.8 & 90.8 & 35.8 & 61.5 \\
 & MGE-PO & 8.9 & 38.3 & 10.2 & 33.3 & 55.3 & 92.5 & 69.4 & 89.2 & 36.0 & 63.3 \\
\addlinespace[3pt]
Qwen2.5-Math-7B\newline  Base & Base & 3.5 & 26.7 & 9.0 & 30.0 & 38.3 & 85.3 & 47.1 & 86.4 & 24.5 & 57.1 \\
 & GRPO & 10.8 & 35.4 & 20.6 & 36.7 & 61.9 & 90.0 & 72.4 & 90.8 & 41.4 & 63.2 \\
 & DAPO & 11.8 & 36.7 & 21.3 & 43.3 & 63.9 & 90.0 & 73.7 & 91.0 & 42.7 & 65.3 \\
 & KL-Cov & 13.1 & 38.3 & 22.9 & 44.4 & 62.4 & 91.3 & 75.8 & 90.6 & 43.6 & 66.2 \\
 & MGE-PO & 13.6 & 43.3 & 23.1 & 50.0 & 64.3 & 95.0 & 73.1 & 91.6 & 43.5 & 70.0 \\
\addlinespace[3pt]
Qwen2.5-7B\newline  Base & Base & 1.3 & 6.7 & 2.3 & 16.7 & 20.0 & 70.0 & 61.8 & 75.4 & 21.3 & 42.2 \\
 & GRPO & 9.6 & 23.3 & 20.8 & 30.3 & 58.7 & 92.5 & 76.7 & 88.2 & 41.5 & 58.6 \\
 & DAPO & 10.1 & 25.7 & 22.1 & 36.7 & 59.5 & 87.5 & 78.3 & 90.4 & 42.5 & 60.1 \\
 & KL-Cov & 12.7 & 28.3 & 22.6 & 39.5 & 61.4 & 91.2 & 80.8 & 89.8 & 44.4 & 62.2 \\
 & MGE-PO & 12.9 & 30.3 & 23.2 & 40.0 & 65.5 & 90.0 & 79.8 & 90.8 & 45.4 & 62.8 \\
\end{longtable}
\endgroup

The original generation limit is approximately 12k tokens. Table 6 reports the separate supplementary protocol; Appendix C identifies the missing repeat and configuration metadata.

\Needspace{120pt}

\subsection{Supplementary Qwen2.5 evaluation and Clip-Cov comparison}

The supplementary protocol evaluates Qwen2.5-7B-Base with 64 samples per question for accuracy and 256 samples for AIME coverage. It is reported separately from the original avg@32/pass@32 protocol and the six-benchmark Qwen3 protocol.

\begingroup
\fontsize{10}{12}\selectfont
\setlength{\tabcolsep}{4pt}
\renewcommand{\arraystretch}{1.15}
\setlength{\tablecontentwidth}{\dimexpr\linewidth-48pt\relax}
\begin{longtable}{@{}L{0.12000\tablecontentwidth}C{0.14700\tablecontentwidth}C{0.14600\tablecontentwidth}C{0.14700\tablecontentwidth}C{0.14700\tablecontentwidth}C{0.14600\tablecontentwidth}C{0.14700\tablecontentwidth}@{}}
\caption{Supplementary Qwen2.5-7B-Base evaluation. Accuracy uses 64 samples per question and AIME coverage uses 256. Methods share the initialization, reported training budget, and evaluation protocol. The KL-coefficient distinction is specified in Appendix D.10. All scores use \%.}\label{tab:6}\\
\toprule
\textbf{Method} & \textbf{MATH-500\newline  avg@64 (\%)} & \textbf{AMC23\newline  avg@64 (\%)} & \textbf{Minerva\newline  avg@64 (\%)} & \textbf{Olympiad\newline  avg@64 (\%)} & \textbf{AIME24\newline  pass@256 (\%)} & \textbf{AIME25\newline  pass@256 (\%)} \\
\midrule
\endfirsthead
\multicolumn{7}{l}{\small\itshape Table 6 (continued)}\\
\toprule
\textbf{Method} & \textbf{MATH-500\newline  avg@64 (\%)} & \textbf{AMC23\newline  avg@64 (\%)} & \textbf{Minerva\newline  avg@64 (\%)} & \textbf{Olympiad\newline  avg@64 (\%)} & \textbf{AIME24\newline  pass@256 (\%)} & \textbf{AIME25\newline  pass@256 (\%)} \\
\midrule
\endhead
\midrule
\multicolumn{7}{r}{\small Continued on next page}\\
\endfoot
\bottomrule
\endlastfoot
GRPO & 66.3 & 50.0 & 35.4 & 37.4 & 36.7 & 53.3 \\
DAPO & 68.3 & \textbf{57.7} & 22.7 & 30.1 & 30.0 & 53.3 \\
KL-Cov & 67.1 & 53.5 & 36.3 & 39.9 & 38.9 & 57.8 \\
MGE-PO & \textbf{68.6} & 55.6 & \textbf{38.5} & \textbf{41.3} & \textbf{40.0} & \textbf{60.0} \\
\end{longtable}
\endgroup

MGE-PO leads five of the six supplementary metrics. On Minerva, its absolute margins over GRPO, DAPO, and KL-Cov are 3.1\%, 15.8\%, and 2.2\%; on Olympiad they are 3.9\%, 11.2\%, and 1.4\%. These 272- and 674-question evaluations broaden the comparison beyond the smaller competition sets. AIME24/AIME25 coverage shows absolute gains of 3.3\%/6.7\% over GRPO and 1.1\%/2.2\% over KL-Cov. DAPO leads AMC23 by an absolute margin of 2.1\%, while MGE-PO's MATH-500 margin over DAPO is 0.3\%.

The available supplementary Clip-Cov comparison covers Minerva and Olympiad under the same 64-sample accuracy protocol:

\begingroup
\fontsize{10}{12}\selectfont
\setlength{\tabcolsep}{4pt}
\renewcommand{\arraystretch}{1.15}
\setlength{\tablecontentwidth}{\dimexpr\linewidth-16pt\relax}
\begin{longtable}{@{}L{0.30000\tablecontentwidth}C{0.35000\tablecontentwidth}C{0.35000\tablecontentwidth}@{}}
\caption{Available supplementary larger-benchmark comparisons. Qwen2.5-7B-Base; 64 responses per question, maximum generation length 8,192 tokens. The available Clip-Cov comparison covers these two benchmarks. All scores use \%.}\label{tab:7}\\
\toprule
\textbf{Method} & \textbf{Minerva-Math avg@64 (\%)} & \textbf{Olympiad avg@64 (\%)} \\
\midrule
\endfirsthead
\multicolumn{3}{l}{\small\itshape Table 7 (continued)}\\
\toprule
\textbf{Method} & \textbf{Minerva-Math avg@64 (\%)} & \textbf{Olympiad avg@64 (\%)} \\
\midrule
\endhead
\midrule
\multicolumn{3}{r}{\small Continued on next page}\\
\endfoot
\bottomrule
\endlastfoot
GRPO & 35.4 & 37.4 \\
DAPO & 22.7 & 30.1 \\
KL-Cov & 36.3 & 39.9 \\
Clip-Cov & 35.1 & 38.4 \\
MGE-PO & \textbf{38.5} & \textbf{41.3} \\
\end{longtable}
\endgroup

\Needspace{120pt}

\subsection{Existing aggregate quadrant diagnostic}

\begingroup
\fontsize{10}{12}\selectfont
\setlength{\tabcolsep}{4pt}
\renewcommand{\arraystretch}{1.15}
\setlength{\tablecontentwidth}{\dimexpr\linewidth-32pt\relax}
\begin{longtable}{@{}L{0.28000\tablecontentwidth}C{0.18000\tablecontentwidth}C{0.18000\tablecontentwidth}C{0.18000\tablecontentwidth}C{0.18000\tablecontentwidth}@{}}
\caption{Aggregate quadrant counts in the original manuscript. Each column summarizes 3,840 rollouts, with a reporting threshold at \(u_i=0.5\). The comparison is initialization versus ordinary GRPO, not GRPO versus MGE-PO.}\label{tab:8}\\
\toprule
\textbf{Diagnostic quadrant} & \textbf{Initialization count} & \textbf{Initialization share (\%)} & \textbf{GRPO count} & \textbf{GRPO share (\%)} \\
\midrule
\endfirsthead
\multicolumn{5}{l}{\small\itshape Table 8 (continued)}\\
\toprule
\textbf{Diagnostic quadrant} & \textbf{Initialization count} & \textbf{Initialization share (\%)} & \textbf{GRPO count} & \textbf{GRPO share (\%)} \\
\midrule
\endhead
\midrule
\multicolumn{5}{r}{\small Continued on next page}\\
\endfoot
\bottomrule
\endlastfoot
Low-entropy correct & 173 & 4.5 & 492 & 12.8 \\
High-entropy correct & 74 & 1.9 & 96 & 2.5 \\
Low-entropy incorrect & 1747 & 45.5 & 1428 & 37.2 \\
High-entropy incorrect & 1846 & 48.1 & 1824 & 47.5 \\
\end{longtable}
\endgroup

Low-entropy incorrect responses account for 45.5\% of initialization rollouts and 37.2\% after GRPO. These aggregate counts establish the coexistence of predictive concentration and incorrectness; they do not identify paired trajectory transitions or MGE-PO error categories.

\Needspace{120pt}

\subsection{Evaluation datasets and protocols}

\begingroup
\fontsize{10}{12}\selectfont
\setlength{\tabcolsep}{4pt}
\renewcommand{\arraystretch}{1.15}
\setlength{\tablecontentwidth}{\dimexpr\linewidth-24pt\relax}
\begin{longtable}{@{}L{0.34000\tablecontentwidth}L{0.10000\tablecontentwidth}L{0.27000\tablecontentwidth}L{0.29000\tablecontentwidth}@{}}
\caption{Qwen2.5 evaluation datasets and reported metrics. Competition subset sizes and the text-only English exact-answer Olympiad subset follow the experimental record. Appendix C lists the reproducibility requirements for exact manifests, grader versions, and overlap filtering.}\label{tab:9}\\
\toprule
\textbf{Benchmark} & \textbf{Questions} & \textbf{Original protocol} & \textbf{Supplementary protocol} \\
\midrule
\endfirsthead
\multicolumn{4}{l}{\small\itshape Table 9 (continued)}\\
\toprule
\textbf{Benchmark} & \textbf{Questions} & \textbf{Original protocol} & \textbf{Supplementary protocol} \\
\midrule
\endhead
\midrule
\multicolumn{4}{r}{\small Continued on next page}\\
\endfoot
\bottomrule
\endlastfoot
MATH-500 & 500 & avg@32, pass@32 & avg@64 \\
AMC 2023 & 40 & avg@32, pass@32 & avg@64 \\
AIME 2024 & 30 & avg@32, pass@32 & pass@256 \\
AIME 2025 & 30 & avg@32, pass@32 & pass@256 \\
Minerva-Math / OCW & 272 & Not evaluated & avg@64 \\
Olympiad, text-only English exact-answer subset & 674 & Not evaluated & avg@64 \\
\end{longtable}
\endgroup

The original Qwen2.5 protocol uses temperature 1.0 and top-\(p\) 0.7. The supplementary Minerva/Olympiad generation limit is 8,192 tokens; Appendix C.2 records the unresolved checkpoint and subset metadata. The separate Qwen3-8B-Base protocol uses avg@32 and pass@32 for all six benchmarks (Table 1; Appendix C.4).

\Needspace{120pt}

\subsection{Instruct initialization}

\begingroup
\fontsize{10}{12}\selectfont
\setlength{\tabcolsep}{4pt}
\renewcommand{\arraystretch}{1.15}
\setlength{\tablecontentwidth}{\dimexpr\linewidth-48pt\relax}
\begin{longtable}{@{}L{0.31000\tablecontentwidth}C{0.11500\tablecontentwidth}C{0.11500\tablecontentwidth}C{0.11500\tablecontentwidth}C{0.11500\tablecontentwidth}C{0.11500\tablecontentwidth}C{0.11500\tablecontentwidth}@{}}
\caption{Qwen2.5-Math-7B-Instruct at the final reported checkpoint. Both methods start from the same Instruct initialization. The first four columns use avg@64; AIME columns use pass@256. The record identifies step 1080 after eight epochs in the setting described as “4k”; the role of this setting is unspecified. All scores and absolute differences use \%.}\label{tab:10}\\
\toprule
\textbf{Method} & \textbf{MATH-500 (\%)} & \textbf{AMC23 (\%)} & \textbf{Minerva (\%)} & \textbf{Olympiad (\%)} & \textbf{AIME24 (\%)} & \textbf{AIME25 (\%)} \\
\midrule
\endfirsthead
\multicolumn{7}{l}{\small\itshape Table 10 (continued)}\\
\toprule
\textbf{Method} & \textbf{MATH-500 (\%)} & \textbf{AMC23 (\%)} & \textbf{Minerva (\%)} & \textbf{Olympiad (\%)} & \textbf{AIME24 (\%)} & \textbf{AIME25 (\%)} \\
\midrule
\endhead
\midrule
\multicolumn{7}{r}{\small Continued on next page}\\
\endfoot
\bottomrule
\endlastfoot
GRPO & 80.3 & 62.4 & 32.4 & 34.6 & 43.3 & 53.3 \\
MGE-PO & \textbf{81.0} & \textbf{63.9} & \textbf{34.5} & \textbf{36.3} & \textbf{46.7} & \textbf{63.3} \\
Absolute difference: MGE-PO minus GRPO (\%) & +0.7 & +1.5 & +2.1 & +1.7 & +3.4 & +10.0 \\
\end{longtable}
\endgroup

The Instruct comparison yields absolute gains over GRPO of 0.7\%, 1.5\%, 2.1\%, 1.7\%, 3.4\%, and 10.0\% in Table 10's benchmark order, extending the two-method comparison to one instruction-tuned initialization. Pre-RL Instruct scores and Instruct DAPO/KL-Cov comparisons are unavailable.

\Needspace{120pt}

\subsection{Quadrant-coefficient sensitivity}

\begingroup
\fontsize{10}{12}\selectfont
\setlength{\tabcolsep}{4pt}
\renewcommand{\arraystretch}{1.15}
\setlength{\tablecontentwidth}{\dimexpr\linewidth-48pt\relax}
\begin{longtable}{@{}L{0.32000\tablecontentwidth}C{0.08000\tablecontentwidth}C{0.08000\tablecontentwidth}C{0.08000\tablecontentwidth}C{0.08000\tablecontentwidth}C{0.18000\tablecontentwidth}C{0.18000\tablecontentwidth}@{}}
\caption{Quadrant-coefficient ablation on Qwen2.5-Math-7B-Base. AIME24 uses the original avg@32/pass@32 protocol. Coefficients are nonnegative magnitudes in the generalized shaping rule: PLR/PHR terms are added and NHR/NLR terms are subtracted. Test-set selection scope is discussed below. Accuracy and coverage use \%; quadrant coefficients are unitless.}\label{tab:11}\\
\toprule
\textbf{Configuration} & \textbf{PLR} & \textbf{NHR} & \textbf{PHR} & \textbf{NLR} & \textbf{avg@32 (\%)} & \textbf{pass@32 (\%)} \\
\midrule
\endfirsthead
\multicolumn{7}{l}{\small\itshape Table 11 (continued)}\\
\toprule
\textbf{Configuration} & \textbf{PLR} & \textbf{NHR} & \textbf{PHR} & \textbf{NLR} & \textbf{avg@32 (\%)} & \textbf{pass@32 (\%)} \\
\midrule
\endhead
\midrule
\multicolumn{7}{r}{\small Continued on next page}\\
\endfoot
\bottomrule
\endlastfoot
Default & 0.5 & 0.3 & 0 & 0 & \textbf{23.1} & \textbf{50.0} \\
PLR-weighted term removed & 0 & 0.3 & 0 & 0 & 21.3 & 36.7 \\
NHR-weighted term removed & 0.5 & 0 & 0 & 0 & 22.2 & 36.7 \\
Magnitudes swapped & 0.3 & 0.5 & 0 & 0 & 22.6 & 40.0 \\
Additional PHR term & 0.5 & 0.3 & 0.3 & 0 & 21.8 & 36.7 \\
NHR-weighted term replaced by NLR-weighted term & 0.5 & 0 & 0 & 0.3 & 21.6 & 33.3 \\
Additional NLR term & 0.5 & 0.3 & 0 & 0.1 & 22.4 & 36.7 \\
\end{longtable}
\endgroup

The quadrant sensitivity study tests the two continuous correction terms and their relative strengths (Table 11). On Math-7B AIME24, removing either default term lowers pass@32 from 50.0\% to 36.7\%; swapping their magnitudes gives 40.0\%. The default \((\gamma_{\mathrm{PLR}},\gamma_{\mathrm{NHR}},\gamma_{\mathrm{PHR}},\gamma_{\mathrm{NLR}})=(0.5,0.3,0,0)\) performs best among the seven tested settings, and is shared across the three base models. The tested additional PHR/NLR terms also reduce performance. This comparison measures local sensitivity on AIME24, which is also used for final evaluation; an independent development-set selection procedure is not established in the experimental record.

\Needspace{120pt}

\subsection{Reference regularization and full-method dynamics}

\begingroup
\fontsize{10}{12}\selectfont
\setlength{\tabcolsep}{4pt}
\renewcommand{\arraystretch}{1.15}
\setlength{\tablecontentwidth}{\dimexpr\linewidth-24pt\relax}
\begin{longtable}{@{}L{0.33000\tablecontentwidth}L{0.20000\tablecontentwidth}L{0.20000\tablecontentwidth}L{0.27000\tablecontentwidth}@{}}
\caption{Token-weight and training-dynamics diagnostics. The token analysis covers approximately 1.78 million active PLR tokens over nine actor updates. Entropy and response-length entries summarize the original plain-7B training curves and have a different scope from the token window.}\label{tab:12}\\
\toprule
\textbf{Diagnostic} & \textbf{GRPO / comparison value} & \textbf{MGE-PO / EA-KL value} & \textbf{Scope} \\
\midrule
\endfirsthead
\multicolumn{4}{l}{\small\itshape Table 12 (continued)}\\
\toprule
\textbf{Diagnostic} & \textbf{GRPO / comparison value} & \textbf{MGE-PO / EA-KL value} & \textbf{Scope} \\
\midrule
\endhead
\midrule
\multicolumn{4}{r}{\small Continued on next page}\\
\endfoot
\bottomrule
\endlastfoot
Mean EA-KL weight in lower-entropy PLR token stratum & Not applicable & 1.19 & Active PLR tokens, nine updates \\
Mean EA-KL weight in higher-entropy PLR token stratum & Not applicable & 0.35 & Same token window \\
Lower-/higher-entropy weight ratio & Not applicable & 3.1--3.4 across updates & Same token window \\
Late-training policy entropy (nats/token) & Approximately 0.001 & Approximately 0.06 & Original Qwen2.5-7B-Base training curves \\
Late-training response length (tokens) & Approximately 6,000 & Approximately 4,500 & Original Qwen2.5-7B-Base training curves \\
\end{longtable}
\endgroup

The PLR token window records lower-entropy mean weights of 1.19 and higher-entropy mean weights of 0.35, with an update-wise ratio of 3.1--3.4. These measurements characterize the implemented allocation of reference cost; the entropy-stratum boundaries are unspecified. Full-method curves combine EA-KL, 4Q, sampling, and the global KL coefficient.

Complete removal of reference KL is a separate intervention from disabling adaptive weights. In the reported no-KL comparison (\(\beta=0\)), training-end reference KL rises from approximately 0.5 to 3 nats and all four benchmark scores show absolute decreases of 4--7\%. This supports retaining reference regularization in the tested setting. Detailed per-benchmark values and the score metric are unreported.

\Needspace{120pt}

\subsection{Complete-run timing}

\begingroup
\fontsize{10}{12}\selectfont
\setlength{\tabcolsep}{4pt}
\renewcommand{\arraystretch}{1.15}
\setlength{\tablecontentwidth}{\dimexpr\linewidth-16pt\relax}
\begin{longtable}{@{}L{0.44000\tablecontentwidth}L{0.28000\tablecontentwidth}C{0.28000\tablecontentwidth}@{}}
\caption{End-to-end wall-clock time for complete single runs with matched training budgets. Matching refers to training budget rather than KL coefficients. The timing record identifies a base setting without a separate model-checkpoint or hardware identifier. These are wall-clock measurements; utilization and GPU-hour accounting are unavailable.}\label{tab:13}\\
\toprule
\textbf{Initialization} & \textbf{Method} & \textbf{Wall-clock hours} \\
\midrule
\endfirsthead
\multicolumn{3}{l}{\small\itshape Table 13 (continued)}\\
\toprule
\textbf{Initialization} & \textbf{Method} & \textbf{Wall-clock hours} \\
\midrule
\endhead
\midrule
\multicolumn{3}{r}{\small Continued on next page}\\
\endfoot
\bottomrule
\endlastfoot
Base setting & GRPO & 22.8 \\
Base setting & KL-Cov & 29.6 \\
Base setting & MGE-PO & 22.6 \\
Qwen2.5-Math-7B-Instruct & GRPO & 23.6 \\
Qwen2.5-Math-7B-Instruct & MGE-PO & 22.8 \\
\end{longtable}
\endgroup

The original experimental hardware is two NVIDIA H200 GPUs with 141 GB each. Entropy is cached as one scalar per generated token. Separate memory overhead was not measured and depends on datatype and residency.

\Needspace{120pt}

\subsection{Curriculum concentration}

\begingroup
\fontsize{10}{12}\selectfont
\setlength{\tabcolsep}{4pt}
\renewcommand{\arraystretch}{1.15}
\setlength{\tablecontentwidth}{\dimexpr\linewidth-24pt\relax}
\begin{longtable}{@{}L{0.43000\tablecontentwidth}C{0.18000\tablecontentwidth}C{0.18000\tablecontentwidth}C{0.21000\tablecontentwidth}@{}}
\caption{Curriculum diagnostics in a 21-step window. The window contains 43,008 rollouts and 2,440 unique MGE-PO questions from the 17,398-question training dataset. Relative ESS, mass, and Gini describe the sampling distribution; unique-question coverage distinguishes uniform expectations from observed draws.}\label{tab:14}\\
\toprule
\textbf{Diagnostic} & \textbf{GRPO, uniform} & \textbf{KL-Cov, uniform} & \textbf{MGE-PO, DynZPD} \\
\midrule
\endfirsthead
\multicolumn{4}{l}{\small\itshape Table 14 (continued)}\\
\toprule
\textbf{Diagnostic} & \textbf{GRPO, uniform} & \textbf{KL-Cov, uniform} & \textbf{MGE-PO, DynZPD} \\
\midrule
\endhead
\midrule
\multicolumn{4}{r}{\small Continued on next page}\\
\endfoot
\bottomrule
\endlastfoot
Relative ESS & 100\% & 100\% & 91.5\% \\
Top-10\% question probability mass & 10.0\% & 10.0\% & 18.3\% \\
Sampling-distribution Gini & 0 & 0 & 0.085 \\
Maximum question share (definition unreported) & Not reported & Not reported & 0.4\% \\
In-window unique-question coverage & 14.3\%, expected & 14.3\%, expected & 14.0\%, observed \\
\end{longtable}
\endgroup

For question distribution \(p\), \(\operatorname{ESS}(p)=1/\sum_qp(q)^2\) and relative ESS is \(100\operatorname{ESS}(p)/|\mathcal D|\) in \%. The averaging convention over the window is unspecified. Appendix C.3 explains the unresolved definition of the 0.4\% maximum-share entry. The finite window characterizes observed concentration and coverage in Section 6.4.

\Needspace{120pt}

\subsection{Qwen2.5 component configurations}

\begingroup
\fontsize{10}{12}\selectfont
\setlength{\tabcolsep}{4pt}
\renewcommand{\arraystretch}{1.15}
\setlength{\tablecontentwidth}{\dimexpr\linewidth-40pt\relax}
\begin{longtable}{@{}L{0.13000\tablecontentwidth}L{0.13000\tablecontentwidth}L{0.15000\tablecontentwidth}C{0.20000\tablecontentwidth}C{0.18000\tablecontentwidth}C{0.21000\tablecontentwidth}@{}}
\caption{Component configurations on Qwen2.5-7B-Base. The score is four-benchmark macro avg@32. EA-KL off retains uniform reference KL.}\label{tab:15}\\
\toprule
\textbf{EA-Adv} & \textbf{EA-KL} & \textbf{DynZPD} & \textbf{Reference coefficient \(\beta\)} & \textbf{Macro avg@32 (\%)} & \textbf{Absolute difference from GRPO (\%)} \\
\midrule
\endfirsthead
\multicolumn{6}{l}{\small\itshape Table 15 (continued)}\\
\toprule
\textbf{EA-Adv} & \textbf{EA-KL} & \textbf{DynZPD} & \textbf{Reference coefficient \(\beta\)} & \textbf{Macro avg@32 (\%)} & \textbf{Absolute difference from GRPO (\%)} \\
\midrule
\endhead
\midrule
\multicolumn{6}{r}{\small Continued on next page}\\
\endfoot
\bottomrule
\endlastfoot
Off & Off & Off & 0.04 & 41.5 & 0.0 \\
On & Off & Off & 0.001 & 43.3 & +1.8 \\
Off & On & Off & 0.001 & 42.2 & +0.7 \\
Off & Off & On & 0.001 & 41.6 & +0.1 \\
On & On & Off & 0.001 & 44.3 & +2.8 \\
On & Off & On & 0.001 & 43.9 & +2.4 \\
Off & On & On & 0.001 & 42.5 & +1.0 \\
On & On & On & 0.001 & \textbf{45.4} & \textbf{+3.9} \\
\end{longtable}
\endgroup

4Q improves all three available shared-coefficient comparisons that add it to a supporting configuration. With \(\beta=10^{-3}\), adding 4Q to EA-KL raises macro avg@32 from 42.2\% to 44.3\% (+2.1\%), and adding it to DynZPD raises the score from 41.6\% to 43.9\% (+2.3\%). Adding 4Q to EA-KL+DynZPD yields 42.5\% to 45.4\% (+2.9\%), the largest conditional gain among removals from the complete configuration. These observed contrasts support outcome-conditioned entropy credit as the central method component.

The complete score also depends on its supporting settings. Adding EA-KL to 4Q+DynZPD gives 43.9\% to 45.4\% (+1.5\%), while adding DynZPD to 4Q+EA-KL gives 44.3\% to 45.4\% (+1.1\%). Thus 45.4\% belongs to the complete implemented configuration. EA-KL off retains uniform reference KL with \(\alpha=0\); on uses \(\alpha=0.20\). The GRPO row uses \(\beta=0.04\), while all seven nonempty configurations use \(0.001\). Full-method differences from GRPO include this coefficient change; the shared-coefficient contrasts report conditional component differences in single training runs. Adaptive weighting can additionally alter realized aggregate penalty strength. Appendix D.7 retains the separate complete-KL-removal observation.

\Needspace{120pt}

\subsection{Qwen3-8B paired 4Q comparisons}

The paired values below supplement Figure 3. All methods use the final controller step 624 and the common evaluation protocol in Appendix C.4.

\begingroup
\fontsize{10}{12}\selectfont
\setlength{\tabcolsep}{4pt}
\renewcommand{\arraystretch}{1.15}
\setlength{\tablecontentwidth}{\dimexpr\linewidth-48pt\relax}
\begin{longtable}{@{}L{0.18000\tablecontentwidth}C{0.13667\tablecontentwidth}C{0.13667\tablecontentwidth}C{0.13667\tablecontentwidth}C{0.13667\tablecontentwidth}C{0.13667\tablecontentwidth}C{0.13667\tablecontentwidth}@{}}
\caption{4Q ablation across three Qwen3-8B-Base optimization settings at the final controller step 624. Each row compares a method with its 4Q-augmented configuration in Table 1. The GRPO comparison uses the row labeled 4Q. Scores are six-benchmark macro averages (\%); differences are absolute score changes (\%) computed before rounding the macro averages to two decimals.}\label{tab:16}\\
\toprule
\textbf{Optimization setting} & \textbf{avg@32\newline  without 4Q (\%)} & \textbf{avg@32\newline  with 4Q (\%)} & \textbf{Absolute \(\Delta\) avg@32 (\%)} & \textbf{pass@32\newline  without 4Q (\%)} & \textbf{pass@32\newline  with 4Q (\%)} & \textbf{Absolute \(\Delta\) pass@32 (\%)} \\
\midrule
\endfirsthead
\multicolumn{7}{l}{\small\itshape Table 16 (continued)}\\
\toprule
\textbf{Optimization setting} & \textbf{avg@32\newline  without 4Q (\%)} & \textbf{avg@32\newline  with 4Q (\%)} & \textbf{Absolute \(\Delta\) avg@32 (\%)} & \textbf{pass@32\newline  without 4Q (\%)} & \textbf{pass@32\newline  with 4Q (\%)} & \textbf{Absolute \(\Delta\) pass@32 (\%)} \\
\midrule
\endhead
\midrule
\multicolumn{7}{r}{\small Continued on next page}\\
\endfoot
\bottomrule
\endlastfoot
GRPO & 41.31 & 45.00 & +3.69 & 57.97 & 69.03 & +11.06 \\
DAPO & 43.61 & 46.99 & +3.38 & 66.38 & 71.69 & +5.31 \\
Clip-Cov & 43.25 & 46.48 & +3.23 & 68.00 & 72.24 & +4.24 \\
\end{longtable}
\endgroup

Both macro metrics improve in all three settings, while individual benchmarks can decrease (Figure 5).

\clearpage

\setlength{\abovedisplayskip}{5pt}\setlength{\belowdisplayskip}{5pt}\setlength{\abovedisplayshortskip}{3pt}\setlength{\belowdisplayshortskip}{3pt}

\section{Formulation Comparison}

\begingroup
\fontsize{10}{12}\selectfont
\setlength{\tabcolsep}{4pt}
\renewcommand{\arraystretch}{1.04}
\setlength{\tablecontentwidth}{\dimexpr\linewidth-32pt\relax}
\begin{longtable}{@{}L{0.13000\tablecontentwidth}L{0.20000\tablecontentwidth}L{0.23000\tablecontentwidth}L{0.23000\tablecontentwidth}L{0.21000\tablecontentwidth}@{}}
\caption{Comparison of intervention rules. \(H_{i,t}\) denotes token predictive entropy, \(h_i^{\mathrm{top}}\) the top-\(K\) pooled rollout statistic, \(u_i\) its decreasing logistic transform, \(r_i\) the binary outcome, \(g_i\) the truncation gate, and \(\xi_q\) the question EMA. \(A_i\) is the base GRPO advantage. 4Q is MGE-PO’s central credit-assignment rule; its curriculum and entropy-weighted reference penalty are included to specify the complete implemented configuration. The MGE-PO advantage expression precedes batch re-centering. An em dash indicates that the cited method introduces no separate intervention in that column.}\label{tab:17}\\
\toprule
\textbf{Method} & \textbf{Signal and granularity} & \textbf{Advantage intervention} & \textbf{KL intervention} & \textbf{Data selection} \\
\midrule
\endfirsthead
\multicolumn{5}{l}{\small\itshape Table 17 (continued)}\\
\toprule
\textbf{Method} & \textbf{Signal and granularity} & \textbf{Advantage intervention} & \textbf{KL intervention} & \textbf{Data selection} \\
\midrule
\endhead
\midrule
\multicolumn{5}{r}{\small Continued on next page}\\
\endfoot
\bottomrule
\endlastfoot
DAPO (Yu et al., 2025) & Group outcomes & Group-relative advantage & Frozen-reference KL omitted & Retains nondegenerate reward groups \\
EntroAdv (Cheng et al., 2026) & Token predictive entropy & Detached entropy bonus, capped by advantage magnitude; Equation (E1) & --- & --- \\
SEED-GRPO (Chen et al., 2025) & Question semantic entropy \(\mathrm{SE}(q)\) & Semantic-entropy attenuation of advantage magnitude; Equation (E2) & --- & --- \\
EAPO v1 (He et al., 2026) & Group-normalized token entropy \(z_{i,t}\) and advantage polarity & Signed, capped token-entropy correction; Equation (E3) & --- & --- \\
KL-Cov (Cui et al., 2025) & Token covariance between log probability and advantage & Base policy-gradient term & \(\beta D_{\mathrm{KL}}(\pi_{\mathrm{old}}\Vert\pi_\theta)\) on top-covariance tokens & --- \\
SENT (Cao et al., 2025) & Question semantic entropy; token entropy and covariance & Base GRPO advantage & Piecewise coefficients: zero on high-entropy tokens; larger on high-covariance low-entropy tokens & Initial-policy semantic-entropy sorting, easy-to-hard stages \\
EDCO (Pang et al., 2026) & Prefix-based inference-entropy estimate & Underlying SFT or RL objective & Underlying training objective & Periodic top-\(N\) high-entropy subset selection \\
MGE-PO & Rollout top-\(K\) entropy; relative token entropy; question EMA & Outcome-conditioned rollout correction; Equation (E4) & Continuous entropy weights multiply reference KL; Equation (E5) & Online intermediate-entropy sampling with a positive floor; Equation (E6) \\
\end{longtable}
\endgroup

\textbf{Advantage and allocation formulas in Table 17.}

The following expressions summarize the interventions compared in Table 17.

\begin{equation}
\adjustbox{max width=\dimexpr\linewidth-36pt\relax}{$\displaystyle A'_t=A_t+\min\!\left(\alpha H_t^{\mathrm{detach}},\lvert A_t\rvert/\kappa\right),\qquad \kappa>1.$}
\tag{E1}\end{equation}

\begin{equation}
\adjustbox{max width=\dimexpr\linewidth-36pt\relax}{$\displaystyle A'_i=A_i f\!\left(\alpha\mathrm{SE}(q)/\mathrm{SE}_{\max}(q)\right).$}
\tag{E2}\end{equation}

In Equation (E2), \(f\) is the nonnegative semantic-entropy attenuation function and \(\mathrm{SE}_{\max}(q)=\log G\) for \(G\) sampled responses to question \(q\). Chen et al.~(2025) consider linear, exponential, and focal attenuation functions.

\begin{equation}
\adjustbox{max width=\dimexpr\linewidth-36pt\relax}{$\displaystyle A'_{i,t}=A_{i,t}+\alpha\operatorname{sign}(A_{i,t})\operatorname{detach}\!\left[
\operatorname{clip}\!\left(z_{i,t},-\lvert A_{i,t}\rvert/\varphi,\lvert A_{i,t}\rvert/\varphi\right)\right].$}
\tag{E3}\end{equation}

\begin{equation}
\adjustbox{max width=\dimexpr\linewidth-36pt\relax}{$\displaystyle A'_i=A_i+g_i\left[r_i\gamma_{\mathrm{PLR}}u_i-(1-r_i)\gamma_{\mathrm{NHR}}(1-u_i)\right].$}
\tag{E4}\end{equation}

\begin{equation}
\adjustbox{max width=\dimexpr\linewidth-36pt\relax}{$\displaystyle w_{i,t}=\operatorname{clip}\!\left(1+\alpha(1-H_{i,t}/\bar H),0,w_{\max}\right).$}
\tag{E5}\end{equation}

\begin{equation}
\adjustbox{max width=\dimexpr\linewidth-36pt\relax}{$\displaystyle P_t(q)\propto\eta+(1-\eta)\exp\!\left[-(\xi_q-\mu)^2/(2\sigma^2)\right].$}
\tag{E6}\end{equation}

The EAPO formulation in this comparison is pinned to arXiv:2604.11056v1. The second version uses the title \emph{Where Hindsight Credit Can Reside: A Signed-Capacity View of Token Updates in RLVR} and the method name HAPO.

\clearpage

\section{Supplementary Figures}

\Needspace{120pt}

\subsection{Per-benchmark 4Q effects on Qwen3-8B-Base}

Figure 5 expands the macro comparison in Figure 3 into its six benchmark contributions. Adding 4Q to GRPO improves both metrics on all six benchmarks. The DAPO and Clip-Cov comparisons include local decreases alongside their positive macro gains.

\begin{figure}[H]
\centering
\includegraphics[width=\linewidth]{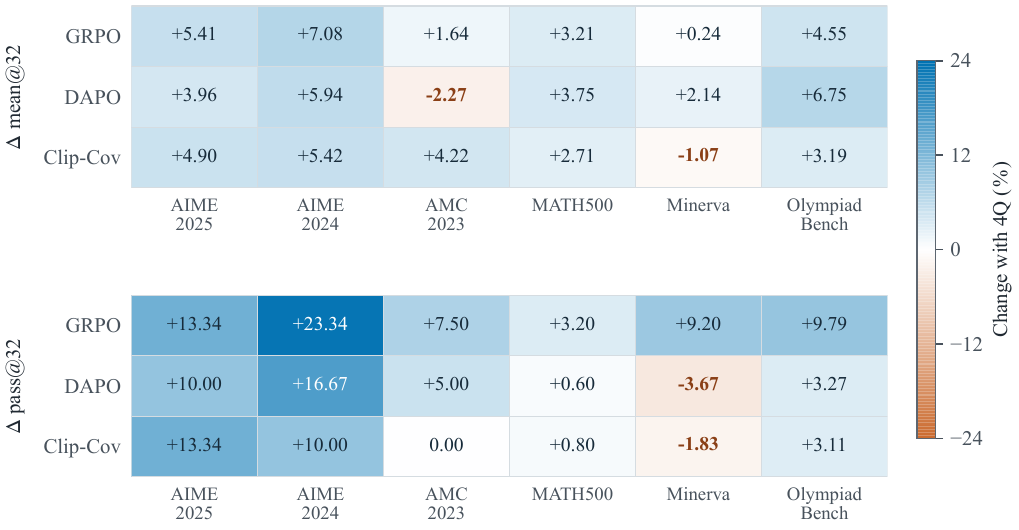}
\caption{Per-benchmark effect of adding 4Q on Qwen3-8B-Base at the final controller step 624. Each cell gives the augmented score minus its baseline score as an absolute difference (\%), for mean@32 (top) and pass@32 (bottom). Rows identify GRPO, DAPO, and Clip-Cov; the corresponding augmented configurations are 4Q, 4Q + DAPO, and Clip-Cov + 4Q. All 36 contrasts share a zero-centered color scale, including negative and zero changes. Differences use the reported two-decimal benchmark scores. Evaluation samples 32 responses per question at temperature 0.6 on identical test sets. Cells describe point estimates without significance claims.}\label{fig:5}
\end{figure}

\Needspace{280pt}

\subsection{Original entropy-stratified comparison}

Figure 6 groups questions by the initial model's generation entropy and compares GRPO with the complete MGE-PO configuration. The highest-entropy bucket has lower reported scores under MGE-PO. These are full-configuration differences across entropy strata.

\begin{figure}[H]
\centering
\includegraphics[width=\linewidth]{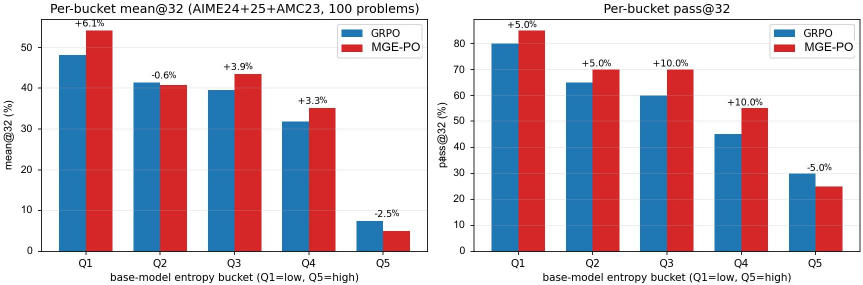}
\caption{Original scores stratified by initial-model entropy. The pooled AIME24, AIME25, and AMC23 set contains 100 questions, divided into five equal-frequency buckets of 20 using top-\(K\) pooled entropy from 64 base-model completions per question (pooling fraction 0.1). Paired bars show GRPO and complete MGE-PO mean@32 (left) and pass@32 (right), both in \%; annotations give absolute score differences (\%). The source does not specify the backbone. This is a full-configuration diagnostic and is separate from the Qwen3-8B 4Q comparisons.}\label{fig:6}
\end{figure}

\Needspace{600pt}

\subsection{Original full-configuration training trajectories}

Figure 7 retains the original Qwen2.5-7B-Base training curves. Late entropy is approximately 0.06 nats/token for MGE-PO versus 0.001 for GRPO, and response length approximately 4,500 versus 6,000 tokens. The early pass@32 advantage contracts later in training. The plotted endpoints differ from Table 2's summary scores; their checkpoint correspondence is undocumented. These curves describe complete configurations and do not isolate 4Q. Figure 4 presents the available shared-coefficient conditional comparisons.

\begin{figure}[H]
\centering
\includegraphics[width=\linewidth]{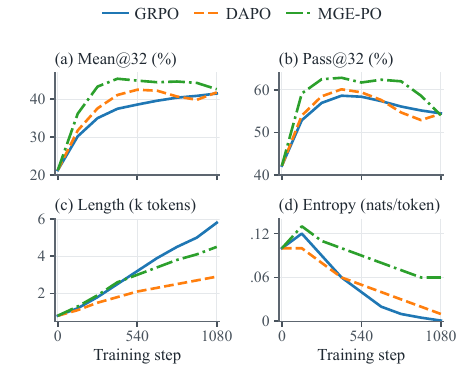}
\caption{Original complete-configuration training trajectories on Qwen2.5-7B-Base over 1080 steps. Panels show four-benchmark mean@32 (avg@32), pass@32, response length, and token entropy for GRPO, DAPO, and MGE-PO. These trajectories are separate from Table 2’s summary scores; their checkpoint correspondence is undocumented. The curves are retained as supplementary diagnostics of complete configurations, not as a 4Q component comparison.}\label{fig:7}
\end{figure}

\end{document}